\def\year{2021}\relax
\documentclass[letterpaper]{article} 
\usepackage{aaai21}  
\usepackage{times}  
\usepackage{helvet} 
\usepackage{courier}  
\usepackage[hyphens]{url}  
\usepackage{graphicx} 
\usepackage{natbib}  
\usepackage{caption} 
\usepackage{amssymb}
\usepackage{amsmath}
\usepackage{algorithm, algorithmic}
\usepackage{dblfloatfix}
\usepackage{subcaption}
\usepackage{booktabs}
\usepackage[switch]{lineno}

\title{Attention Option-Critic}
\author{

    Raviteja Chunduru\textsuperscript{\rm 1,2}, 
    Doina Precup\textsuperscript{\rm 1,2,3}
}
\affiliations{

    \textsuperscript{\rm 1}McGill University, 
    \textsuperscript{\rm 2}Mila, 
    \textsuperscript{\rm 3}Google Deepmind

    raviteja.chunduru@mail.mcgill.ca, dprecup@cs.mcgill.ca

}

\begin{document}


\maketitle

\begin{abstract}
Temporal abstraction in reinforcement learning is the ability of an agent to learn and use high-level behaviors, called options. The option-critic architecture provides a gradient-based end-to-end learning method to construct options. We propose an attention-based extension to this framework, which enables the agent to learn to focus different options on different aspects of the observation space. We show that this leads to behaviorally diverse options which are also capable of state abstraction, and prevents the degeneracy problems of option domination and frequent option switching that occur in option-critic, while achieving a similar sample complexity. We also demonstrate the more efficient, interpretable, and reusable nature of the learned options in comparison with option-critic, through different transfer learning tasks. Experimental results in a relatively simple four-rooms environment and the more complex ALE (Arcade Learning Environment) showcase the efficacy of our approach.
\end{abstract}

\section{Introduction}
\label{intro}

Humans are effortlessly adept at many forms of abstraction. We use a high-level of decision making, to plan using abstract actions, which are typically composed of a sequence of shorter, lower-level actions and last for an extended period of time. This is known as temporal abstraction. When observing our surroundings before making a decision, we also rely and focus on  the important aspects of our sensory input, and ignore  unnecessary signals. This is called state abstraction. 

Within the options framework \cite{Sutton99,Precup00}, for temporal abstraction, the end-to-end learning of hierarchical behaviors has recently become possible through the option-critic architecture \cite{Bacon17}, which enables the learning of intra-option policies, termination functions and the policy over options, to maximize the expected return. However, if this is the sole objective for option discovery, the benefit that the learned options have over primitive action policies becomes questionable. Indeed, the option-critic architecture eventually results in option degeneracy i.e. either one option dominates and is the only one that is used, or there is frequent termination of and switching between options. Introduced to combat this problem, the deliberation cost model \cite{Harb18} modifies the termination gradient to assign a penalty to option termination. This leads to extended options, but relies on  a hard-to-interpret cost parameter. Alternatively, the termination critic algorithm \cite{Harutyunyan19} employs a predictability objective for option termination to prevent option collapse and improve planning. 

In this paper, we adopt the view that options should be diverse in their behavior by explicitly learning to attend to different parts of the observation space. This approach would solve the degeneracy problem by ensuring that options are only used when their respective attentions are activated. This is in line with the notion of options specializing to achieve specific behaviors, and can be viewed as an approach that simultaneously achieves state and temoral abstraction. 
Our approach, in effect, relaxes the strong assumption -- made by most option discovery methods -- that all options are available everywhere, and is a step towards learning the initiation sets for options \cite{Sutton99}, which are otherwise inconvenient to directly learn using a gradient-based  approach. We present an algorithm which builds on option-critic in order to develop options endowed with attention, and evaluate it in several problems, showing both quantitative benefits, as well as qualitative benefits, in terms of the interpretability of the options that are acquired.

The view of bounded rationality \cite{Simon57} can be seen as one of the motivations for temporal abstraction. The added capability of state abstraction takes this one step further, and serves as an additional rationale for our work.

\section{Background}
\label{background}

A discrete-time finite discounted MDP (Markov Decision Process) $\mathcal{M}$ \cite{Puterman95,Barto98} is defined as a  tuple ${\{\mathcal {S},\mathcal{A},R,P,\gamma\}}$, where $\mathcal{S}$ is the set of states, $\mathcal{A}$ is the set of actions, $R : \mathcal{S} \times \mathcal{A} \rightarrow \mathbb{R} $ is the reward function, $P : \mathcal{S} \times \mathcal{A} \times \mathcal{S} \rightarrow [0,1]$ is the transition probability function which specifies the dynamics of the environment, and $\gamma \in [0,1)$ is the scalar discount factor. A Markovian stationary policy $\pi : \mathcal{S} \times \mathcal{A} \rightarrow [0,1] $ is a probabilistic mapping from the set of states $\mathcal{S}$ to the set of actions $\mathcal{A}$. At each timestep $t$, the agent observes state $s_{t} \in \mathcal{S}$ and takes an action $a_{t} \in \mathcal{A}$ according to policy $\pi$, thereby receiving reward $r_{t+1} = R(s_{t},a_{t})$ and transitioning to state $s_{t+1} \in \mathcal{S}$ with probability $P(s_{t+1}|s_t,a_t)$. For policy $\pi$, the discounted state value function is given by: $V^{\pi}(s) = \mathbb{E}_{\pi}[\sum_{t=0}^{\infty} \gamma^{t}r_{t+1} | s_0=s]$ and the discounted action value function by: $Q^{\pi}(s,a) = \mathbb{E}_{\pi}[\sum_{t=0}^{\infty} \gamma^{t}r_{t+1} | s_0=s,a_0=a]$.

For a parameterized policy $\pi_\theta$ with $J(\theta)=V^\pi(s_0)$ as the objective function (where $s_0$ is an initial state), the policy gradient theorem \cite{Sutton00} can be used to derive a learning algorithm for finding the optimal policy $\pi_\theta^*$ that maximizes $J(\theta)$ as:
\begin{equation}
\nabla_{\theta}J(\theta, s_0) = \sum_{s}d^{\pi}(s|s_0)\sum_{a}\nabla_{\theta}\pi(s,a)Q^{\pi}(s,a)    
\end{equation}
where $d^\pi(s|s_0) = \sum_{t=0}^{\infty}\gamma^{t}P(s_t = s|s_0, \pi)$ is the discounted weighting of states with $s_0$ as the starting state. In actor-critic methods \cite{Konda00}, the action values $Q^{\pi}_{\phi}(s,a)$, parameterized by $\phi$, are typically estimated by using temporal difference (TD) learning \cite{Sutton88}. For instance, the update rule for 1-step TD(0) is: $\phi = \phi + \alpha\delta_{t}\nabla_{\phi}Q^{\pi}_{\phi}(s_t,a_t)$ where the TD(0) error $\delta_t = r_{t+1}+\gamma Q^{\pi}_{\phi}(s_{t+1},a_{t+1}) - Q^{\pi}_{\phi}(s_t,a_t)$ and $\alpha\in (0,1)$ is the learning rate. 

\subsection{The Options Framework}
A Markovian option $\omega \in \Omega$ \cite{Sutton99} is a tuple that consists of an initiation set $\mathcal{I}_\omega \subseteq \mathcal{S}$, which denotes the  set of states where the option can be initiated, an intra-option policy $\pi_\omega : \mathcal{S} \times \mathcal{A} \rightarrow [0,1]$, which specifies a probabilistic mapping from states to actions, and a termination condition $\beta_\omega : \mathcal{S} \rightarrow [0,1]$, which signifies the probability of option termination in any state. Let $\Omega(s)$ denote the set of available options for state $s$. An option $\omega$ is available in state $s$ if $s \in \mathcal{I}_\omega$. The set of all options  is: $\Omega=\cup_{s\in \mathcal{S}} \Omega(s)$.

Similar to \citet{Bacon17}, we consider call-and-return option execution. In this model, when the agent is in state $s_t$, it chooses an option $\omega \in \Omega(s_t)$ according to a policy over options $\pi_\Omega$. The intra-option policy $\pi_\omega$ is then followed until the current option terminates according to $\beta_\omega$ after which a new option that is available at the new state is chosen by $\pi_\Omega$, and the process repeats. Like many existing option discovery methods, we too make the assumption that all options are available everywhere, i.e., $\forall s\in\mathcal{S}, \forall \omega \in \Omega : s \in \mathcal{I}_\omega$. However, we show that our approach ends up relaxing this assumption, in effect, and provides an elegant way to learn distinct initiation sets for options. 

The option-critic architecture \cite{Bacon17} provides an end-to-end gradient-based method to learn options. For parameterized intra-option policies $\pi_{\omega,\theta}$ and option terminations $\beta_{\omega,\nu}$, the option-value function is:
\begin{equation}
Q_{\Omega}(s,\omega) = \sum_{a}\pi_{\omega,\theta}(a | s)Q_{U}(s,\omega,a)   
\end{equation}
where $Q_{U} : \mathcal{S} \times \Omega \times \mathcal{A} \rightarrow \mathbb{R}$ is the value of executing action $a$ in the context of state-option ($s,\omega$):
\begin{equation}
Q_{U}(s,\omega,a) = r(s,a) + \gamma\sum_{s'}P(s' | s,a)U(\omega,s')   
\end{equation}
and $U : \Omega \times \mathcal{S} \rightarrow \mathbb{R}$ is the option-value on arrival \cite{Sutton99} and represents the value of executing option $\omega$ in state $s'$:
\begin{equation}
U(\omega,s') = (1-\beta_{\omega,\nu}(s'))Q_{\Omega}(s',\omega) + \beta_{\omega,\nu}(s')V_{\Omega}(s')   
\end{equation}
where $V_{\Omega}(s) = \sum_{\omega}\pi_{\Omega}(\omega | s)Q_{\Omega}(s,\omega)$ is the option-level state value function. 

The intra-option policies and option terminations can be learned by using the policy gradient theorem to maximize the expected discounted return \cite{Bacon17}. The gradient of this objective with respect to intra-option policy parameters $\theta$ when the initial condition is $(s_0,\omega_0)$ is:
\begin{equation}
\begin{aligned}
\nabla_{\theta}J(\theta, s_0, \omega_0) =  \sum_{s,\omega}\big\{\mu_{\Omega}(s,\omega|s_0,\omega_0)\\ \times \sum_{a}\big[\nabla_{\theta}\pi_{\omega,\theta}(a|s)\big]Q_{U}(s,\omega,a)\big\}
\end{aligned}
\end{equation}
where $\mu_{\Omega}(s,\omega|s_0,\omega_0)$ is the discounted weighting of state-option pairs along trajectories that start with $(s_0,\omega_0)$: $\mu_{\Omega}(s,\omega|s_0,\omega_0) = \sum_{t=0}^{\infty}\gamma^{t}P(s_t=s,\omega_t=\omega|s_0,\omega_0)$. Similarly, the gradient with respect to option termination parameters $\nu$ with initial condition $(s_1,\omega_0)$ is:
\begin{equation}
\begin{aligned}
\nabla_{\nu}J(\nu, s_1, \omega_0) = -\sum_{s',\omega}\big\{\mu_{\Omega}(s',\omega|s_1,\omega_0) \\ \times \big[\nabla_{\nu}\beta_{\omega,\nu}(s')\big]A_{\Omega}(s',\omega)\big\}   
\end{aligned}
\end{equation}
where $A_{\Omega}(s,\omega) = Q_{\Omega}(s,\omega) - V_{\Omega}(s)$ is the advantage of choosing option $\omega$ in state $s$.

\subsection{Attention}
The attention mechanism was first proposed in language translation tasks \cite{Bahdanau15} but has since been applied in vision \cite{Mnih14} and reinforcement learning \cite{Sorokin15} as well. It enables the localization of important information before making a prediction. In our approach, soft attention (smoothly varying and differentiable) is applied as a learnable mask over the state observations.

\section{Attention Option-Critic}
\label{aoc}
One inspiration for using attention with options stems from human behavior. Consider the situation of putting on a coat that is hanging next to your front door, opening the door, and heading out. Consider one option to be putting on your coat and another to be opening the door, with a field of view where both coat and door are visible. When executing the first option, we do not focus on the door even though it is visible. Similarly, when opening the door, we do not consider our coat anymore. The high-level behavior that we exhibit influences what we pay attention to, which in turn determines whether a behavior can and should be executed in that particular situation.   

We introduce the Attention Option-Critic (AOC) architecture in order to enable options to learn to attend to specific features in the observation space, in order to diversify their behavior and prevent degeneracy. An attention mechanism $h_{\omega,\phi}$, parameterized by $\phi$, is applied to the observation $s$ for each option $\omega$ as: $o_{\omega} = h_{\omega,\phi}(s)\odot s$ where $\odot$ denotes element-wise multiplication. $h_{\omega,\phi}$ consists of values in $[0,1]$ and is the same size as the original observation $s$. The result $o_\omega$ is used to determine the value of the option, the intra-option policy and the termination condition. This is done for each option separately, and ensures only the required features from the observation determine the option's behavior. We refer to $o$ as the list of all attention-modified observations for each option $o = \{o_{\omega} : \omega \in \Omega\}$. The learning of the option terminations and intra-option policies is performed similarly to the option-critic architecture. The complete algorithm is shown in Algorithm \ref{alg:aoc}.

\begin{algorithm}[tb]
\caption{Attention Option-Critic}
\label{alg:aoc}
\begin{algorithmic}
   \STATE {\bfseries Input:} $\alpha_{\theta}, \alpha_{\nu}, \alpha_{\phi}$ as learning rates for $\theta, \nu$ and $\phi$ respectively.
   \STATE Initialize policy over options $\pi_{\Omega}(o)$, intra-option policies $\pi_{\omega,\theta}$, option terminations $\beta_{\omega,\nu}$, option attentions $h_{\omega,\phi}$
   \STATE $s \leftarrow s_0$
   \STATE $o \leftarrow \{h_{\omega,\phi}(s)\odot s : \omega \in \Omega\}$
   \STATE Choose $\omega$ according to $\epsilon$-soft $\pi_{\Omega}(o)$
   \REPEAT
   \STATE Choose $a$ according to $\pi_{\omega,\theta}(a|o_{\omega})$
   \STATE Take action $a$ in $s$, observe $s',r$
   \STATE $o' \leftarrow \{h_{\omega,\phi}(s')\odot s' : \omega \in \Omega\}$
   \STATE
   \STATE \bfseries{1. Options evaluation:}
   \STATE $\delta \leftarrow r - Q_{U}(o_{\omega},\omega,a)$
   \IF {$s'$ {\normalfont is non-terminal}}
   \STATE $\delta \leftarrow \delta + \gamma(1-\beta_{\omega,\nu}(o'_{\omega}))Q_{\Omega}(o'_{\omega},\omega) $\\\hspace{2em}$+ \gamma\beta_{\omega,\nu}(o'_{\omega})\max_{\bar{\omega}}Q_{\Omega}(o'_{\omega},\bar{\omega})$
   \ENDIF
   \STATE $Q_{U}(o_{\omega},\omega,a) \leftarrow Q_{U}(o_{\omega},\omega,a) + \alpha\delta$
   \STATE
   \STATE \bfseries{2. Options improvement:}
   \STATE $\theta \leftarrow \theta + \alpha_{\theta}\big[\nabla_{\theta}\log\pi_{\omega,\theta}(a|o_{\omega})\big]Q_{U}(o_{\omega},\omega,a)$
   \STATE $\nu \leftarrow \nu - \alpha_{\nu}\big[\nabla_{\nu}\beta_{\omega,\nu}(o'_{\omega})\big]\big[Q_{\Omega}(o'_{\omega},\omega) - V_{\Omega}(o'_{\omega})\big]$
   \STATE $\phi \leftarrow \phi+\alpha_{\phi}\nabla_{\phi}\big[Q_{\Omega}(o_\omega,\omega)+L\big]$
   \STATE
   \IF {$\beta_{\omega,\nu}$ {\normalfont terminates in} $s'$}
   \STATE {\normalfont choose new $\omega$ according to $\epsilon$-soft $\pi_{\Omega}(o')$}
   \ENDIF
   \STATE $s \leftarrow s'$
   \STATE $o \leftarrow \{h_{\omega,\phi}(s)\odot s : \omega \in \Omega\}$
   \UNTIL{$s'$ is terminal}
\end{algorithmic}
\end{algorithm}

\begin{figure}[!b]
\centering
\includegraphics[scale=0.32]{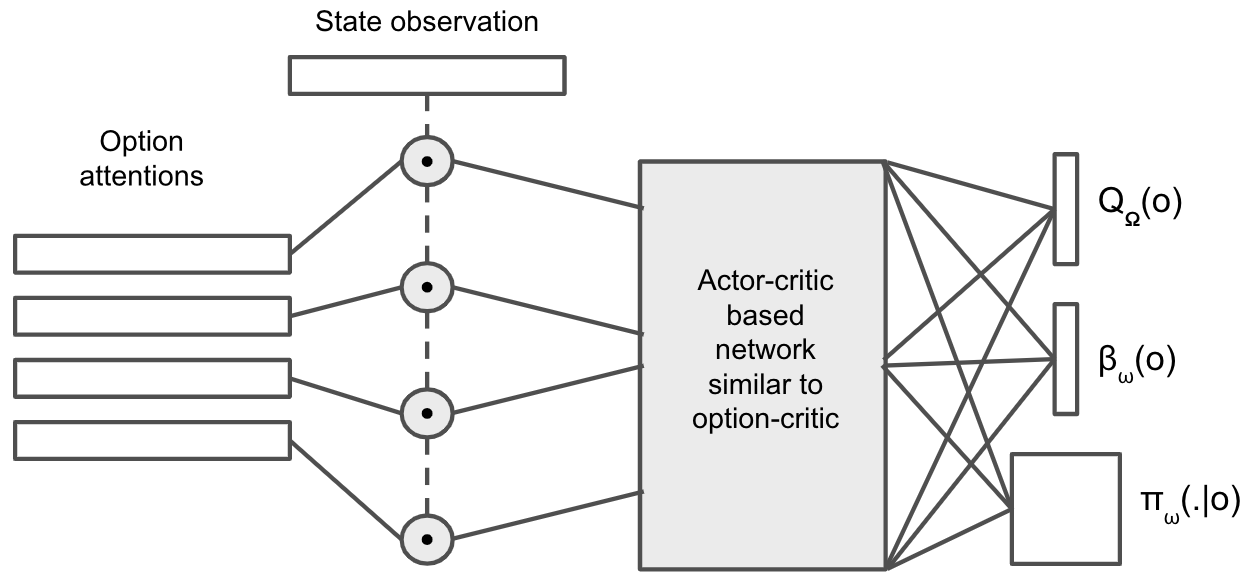}
\caption{The shared network model for AOC in four-rooms. The option attentions are observation independent. The $\odot$ symbol represents element-wise multiplication.}
\label{4rooms_archi}
\end{figure}

The attention for each option is learned to maximize the expected cumulative return of the agent while simultaneously maximizing a distance measure between the attentions of the options, so that the options attend to different features. Additionally, some regularization is added to facilitate the emergence of desired option characteristics. The attention parameters $\phi$ are updated with gradient ascent as $\phi = \phi+\alpha_{\phi}\nabla_{\phi}\big[Q_{\Omega}(o_\omega,\omega)+L\big]$, where $L$ denotes the sum of the distance measure and the regularization, weighted by their respective importance. More details are specified in the next section. 

The attention mechanism brings an aspect of explainability to the agent, and allows one to easily understand each option's focus and behavior. Also, it provides a highly interpretable knob to tune options since the characteristics of the resulting options can be controlled by affecting how the attentions of the options are learned during training. For example, constraining attentions to be distinct enables the diversity of options to be set explicitly as a learning objective. Alternatively, penalizing differences in option attention values for states along a trajectory results in temporally extended options,  achieving an effect similar to the deliberation cost model \cite{Harb18}, but in a more transparent way. 

The resulting attention for each option also serves as an indication of the regions of state space where that option is active and can be initialized. Thus, along with the intra-option policies and option terminations, AOC essentially learns the initiation sets of the options, in that an option is typically only initiated in a particular state when the corresponding attention of that option in that state is high. This prevents the options from degenerating. Since every option cannot be executed or initiated everywhere, this  prevents both frequent option termination and switching, and also  option domination (see section \ref{dominant_option_usage_appendix}), by ensuring that a single option cannot always be followed. 

\subsection{Optimality of Learned Solution}

Since each option receives different information, it is not immediately obvious whether the solution that is learned by AOC is flat, hierarchically or recursively optimal \cite{dietterich00}. However, each option learns to act optimally based on the information that it sees, and apart from some constraints enforced via option attentions, individual option optimality is driven by the goal of the policy over options to maximize the total expected return. Since there is no pseudo reward or subtask assigned to each option, their attentions and areas of usage are learned to maximize this objective and we reason that in the absence of attention constraints, a flat optimal policy will be learned in the limit. In the presence of constraints, the optimality of the learned options will depend on the interactions between the multiple objectives. Even in such cases, AOC is capable of achieving a flat optimal solution, as shown empirically in the next section.

\begin{figure*}[!t]
    \centering
    \begin{subfigure}{0.05\linewidth}
      \caption{}
      \label{options_plots_a}
    \end{subfigure}
    \begin{subfigure}{0.54\linewidth}
        \includegraphics[width=\linewidth]{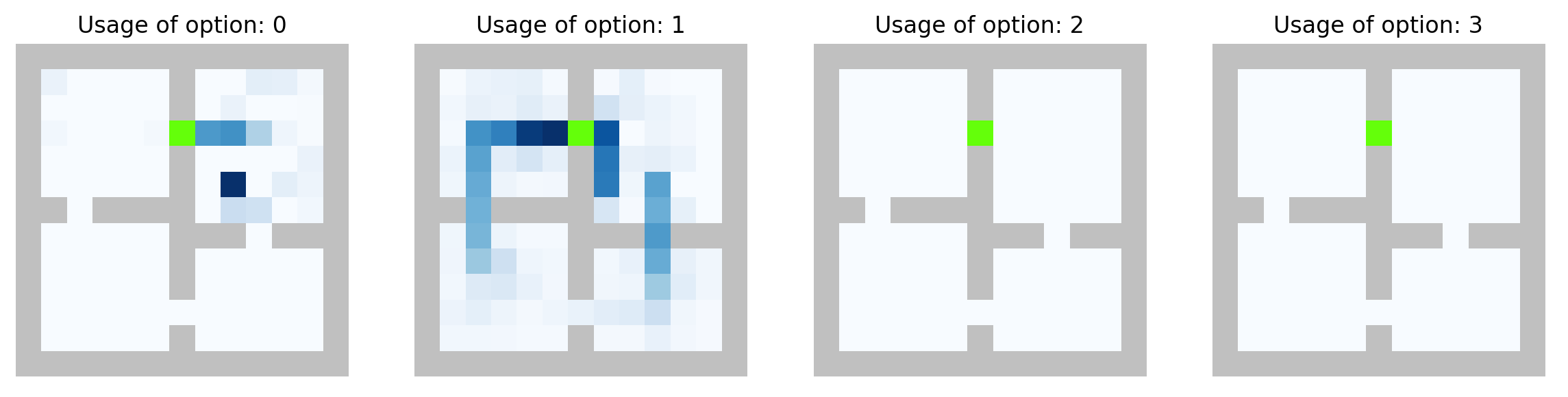}
    \end{subfigure}

    \begin{subfigure}{0.05\linewidth}
      \caption{}
      \label{options_plots_b}
    \end{subfigure}
    \begin{subfigure}{0.54\linewidth}
        \includegraphics[width=\linewidth]{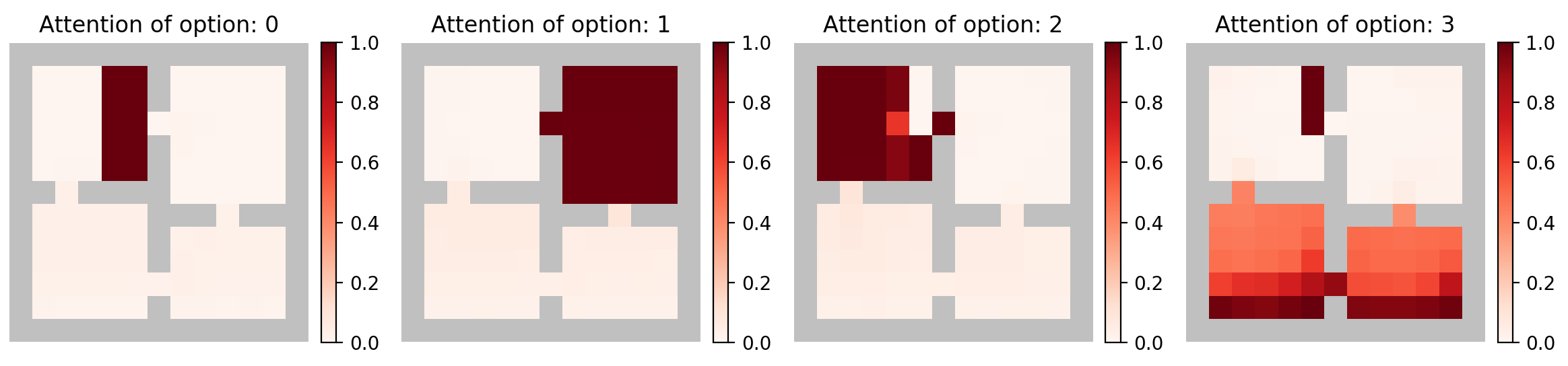}
    \end{subfigure}

    \begin{subfigure}{0.05\linewidth}
      \caption{}
      \label{options_plots_c}
    \end{subfigure}
    \begin{subfigure}{0.54\textwidth}
        \includegraphics[width=\textwidth]{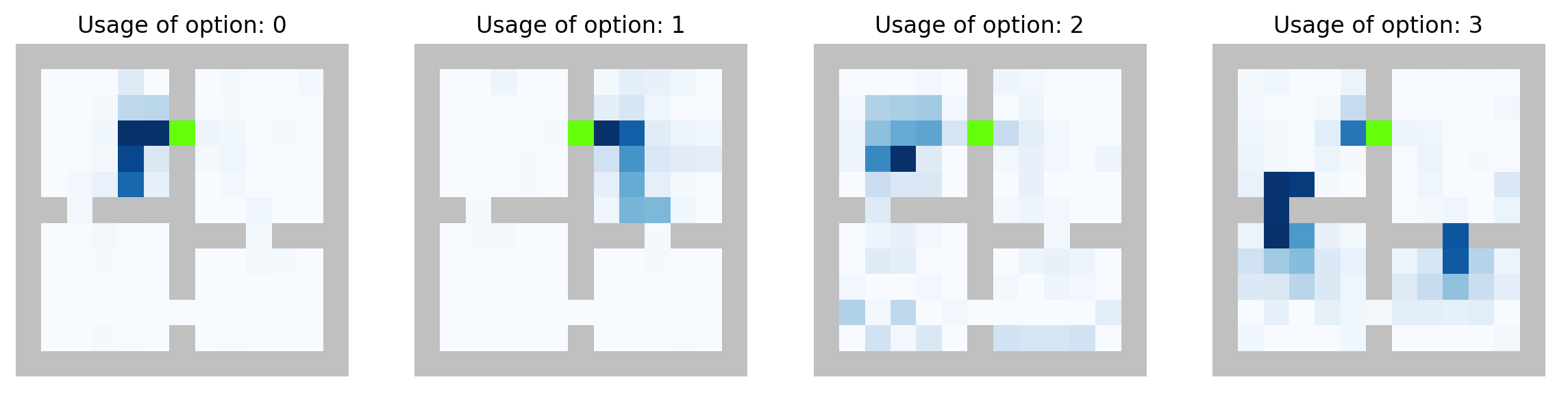}
    \end{subfigure}
    
    \caption{An example of learned options in the four-rooms domain with goal at the north hallway (shown in green). \textbf{(a)} example of degenerate options learned by OC. Darker color indicates more frequent option execution in that particular state. Option 1 dominates and is used 88.12\% while Option 0 is used 11.88\%. Options 2 and 3 are unused. \textbf{(b)} the resulting attention learned for each option with AOC. \textbf{(c)} the options learned using AOC. The options are diverse and respect their attentions. The option usage is relatively balanced at 19.5\%, 34.3\%, 8.1\% and  38.1\% respectively.}
    \label{options_plots}
\end{figure*}

\begin{figure*}[!b]
  \centering
  \subfloat[][Training curves]{\includegraphics[scale=0.4]{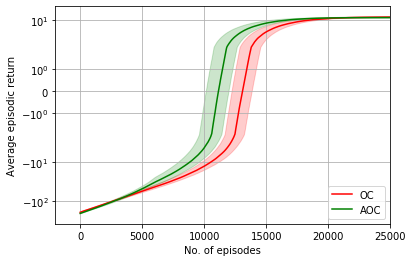} \label{training_curves_a}}
  \subfloat[][Goal transfer]{\includegraphics[scale=0.4]{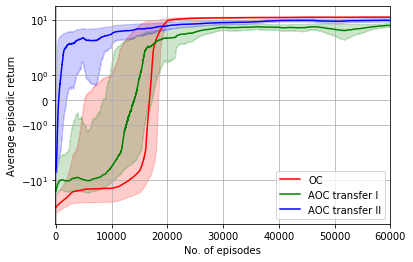}\label{training_curves_b}}
  \subfloat[][Blocked hallway]{\includegraphics[scale=0.4]{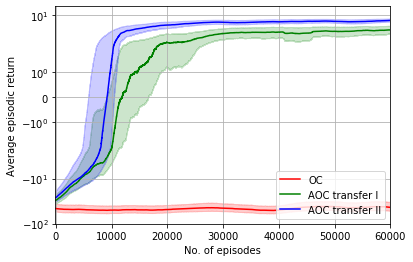}\label{training_curves_c}}
\caption{Learning and transfer (averaged over 10 runs) in the four-rooms domain with 4 options.}
\label{training_curves}
\end{figure*}

\section{Experimental Results}
\label{experiments}

In this section, we show the benefit of combining attention with options and empirically demonstrate that it prevents option degeneracy, provides interpretability, and promotes reusability of options in transfer settings. 

\subsection{Learning in the four-rooms environment}

\begin{figure*}[!t]
    \centering
    \begin{subfigure}{0.05\linewidth}
      \caption{}
      \label{options_transfer_a}
    \end{subfigure}
    \begin{subfigure}{0.54\linewidth}
        \includegraphics[width=\linewidth]{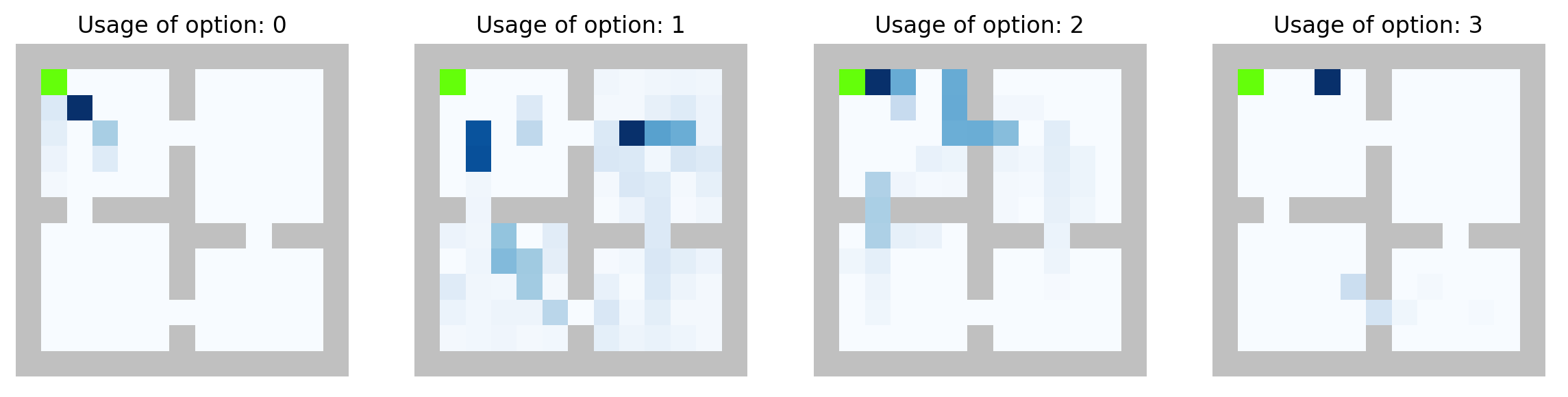}
    \end{subfigure}

    \begin{subfigure}{0.05\linewidth}
      \caption{}
      \label{options_transfer_b}
    \end{subfigure}
    \begin{subfigure}{0.54\linewidth}
        \includegraphics[width=\linewidth]{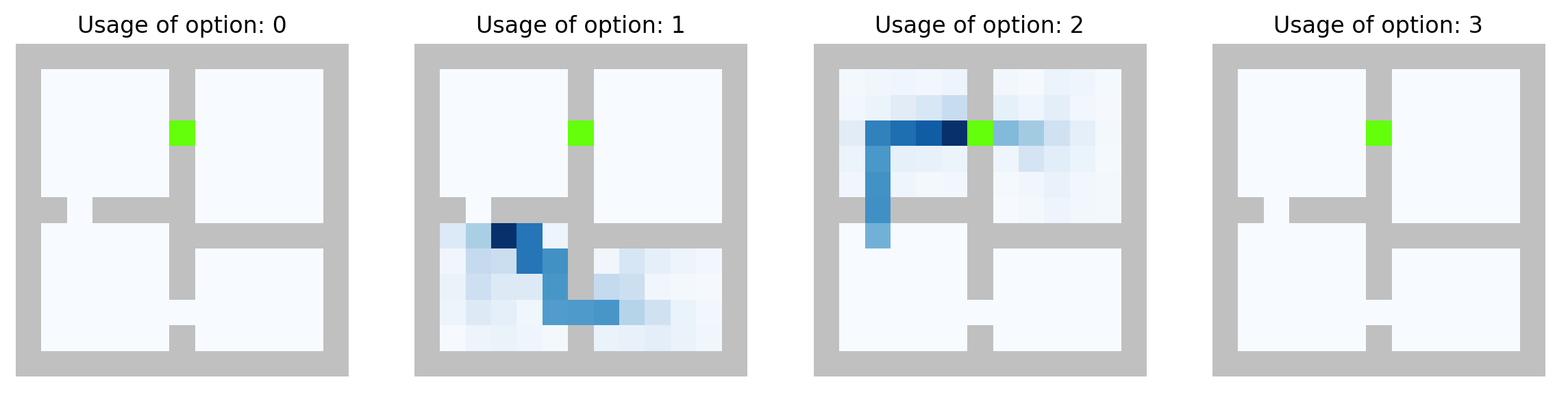}
    \end{subfigure}

    \begin{subfigure}{0.05\linewidth}
      \caption{}
      \label{options_transfer_c}
    \end{subfigure}
    \begin{subfigure}{0.54\textwidth}
        \includegraphics[width=\textwidth]{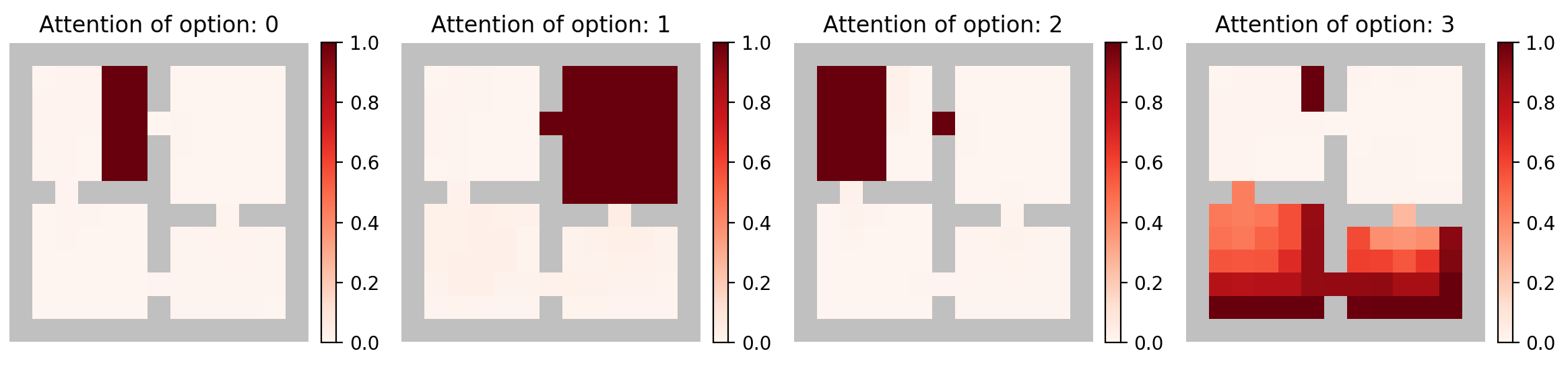}
    \end{subfigure}
    
    \begin{subfigure}{0.05\linewidth}
      \caption{}
      \label{options_transfer_d}
    \end{subfigure}
    \begin{subfigure}{0.54\linewidth}
        \includegraphics[width=\linewidth]{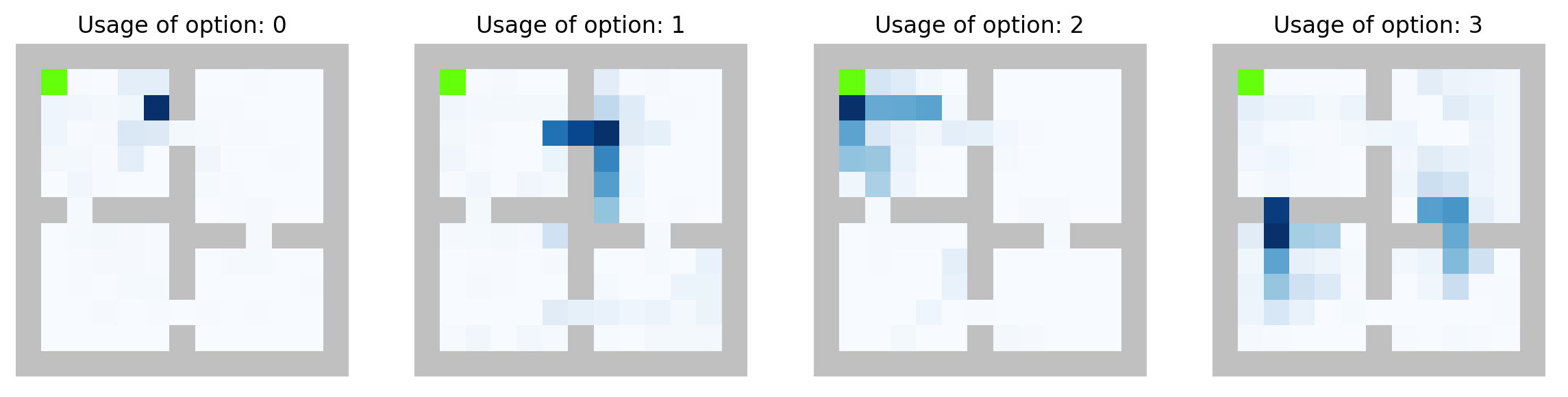}
    \end{subfigure}

    \begin{subfigure}{0.05\linewidth}
      \caption{}
      \label{options_transfer_e}
    \end{subfigure}
    \begin{subfigure}{0.54\linewidth}
        \includegraphics[width=\linewidth]{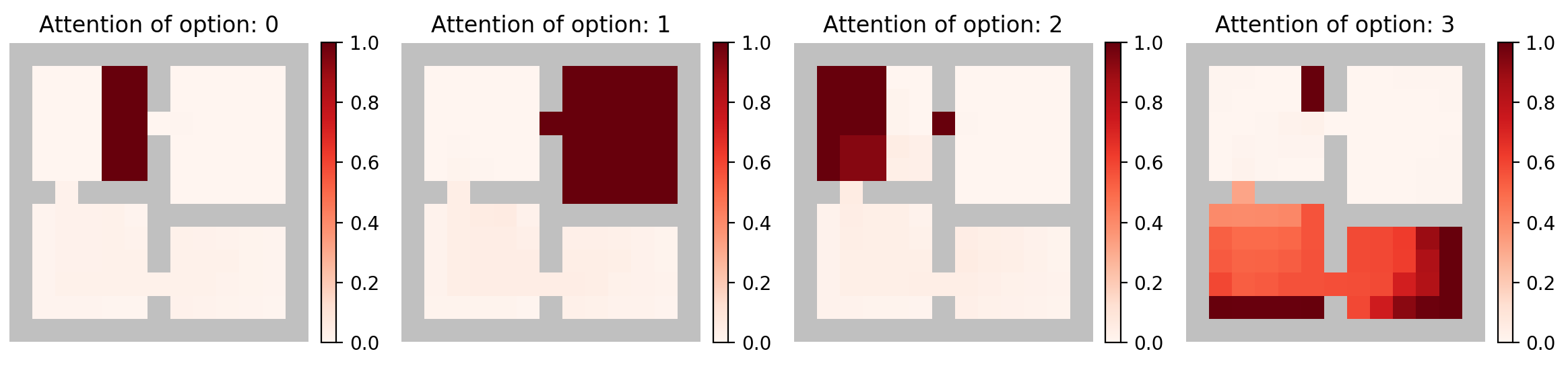}
    \end{subfigure}

    \begin{subfigure}{0.05\linewidth}
      \caption{}
      \label{options_transfer_f}
    \end{subfigure}
    \begin{subfigure}{0.54\textwidth}
        \includegraphics[width=\textwidth]{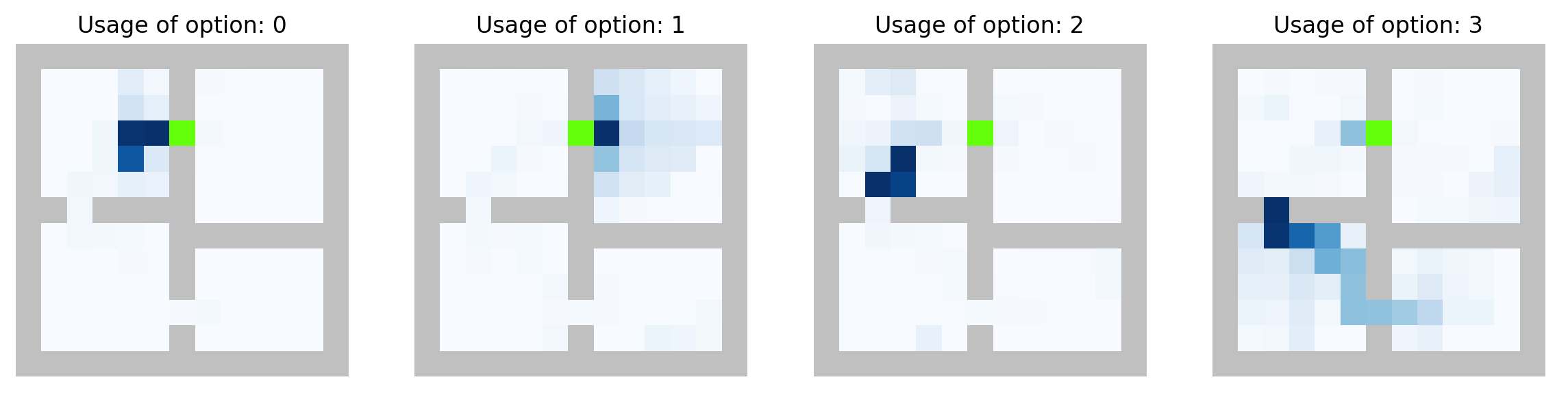}
    \end{subfigure}

    \caption{\textbf{(a) and (b):} resulting options learned by option-critic upon goal transfer and transfer with blocked hallway respectively. \textbf{(c) and (d):} the resulting option attentions and usage upon goal transfer. \textbf{(e) and (f):} the resulting option attentions and usage upon transfer with blocked hallway. For the goal transfer plots above, the goal is shifted from north hallway to the top left state in the north west room. For blocked hallway transfer, the goal is kept fixed as the north hallway, but the east hallway is blocked. The transfer results shown here are with OC and AOC transfer I. The goal states are shown in green.}
    \label{options_transfer}
\end{figure*}

We start by showing the benefit of attention in the four-rooms navigation task \cite{Sutton99} where the agent must reach a specified goal. The observation space consists of one-hot-encoded vectors for every permissible state in the grid. The available actions are up, down, left and right. The chosen action is executed with probability 0.98 and a random action is executed with 0.02 probability. The reward is +20 upon reaching the goal, and -1 otherwise. The agent starts in a uniformly random state and the goal is chosen randomly for each run of the algorithm. 

We use 4 options for learning, with a discount factor of 0.99. The attention $h_{\omega,\phi}$ for each option $\omega$ is initialized randomly as a vector of the same length as the input observation $s$. Thus, in this situation, the option attentions are independent of the state observation. We employ a 3-layer shared-parameter neural network with 3 individual heads, each of which respectively estimates the intra-option policy, the option termination functions, and the option values. In our implementation of AOC (for all experiments), the network learns the option values $Q_\Omega$ to which the $\epsilon$-greedy strategy is applied to determine the policy over options $\pi_\Omega$. Intra-option exploration is enforced with entropy regularization. The architecture is shown in Figure \ref{4rooms_archi}.

The option attentions, option values, intra-option policies and option terminations are learned in an end-to-end manner to maximize the total expected discounted return. The cosine similarity between the option attentions is added to the loss to ensure that the learned attentions are diverse. Furthermore, a regularization loss -- the sum of absolute differences between attentions (for each option) of adjacent states in a trajectory -- penalizes irregularities in the option attentions of states in close temporal proximity along trajectories. This results in smooth attentions and leads to temporally extended options by minimizing switching, and achieves a similar effect to deliberation cost \cite{Harb18}. Thus, for the four-rooms domain, the $L$ term in Algorithm \ref{alg:aoc} is enforced by adding $w_{1}L_{1}+w_{2}L_{2}$ to the overall network loss function, where $L_1$ is the total sum of cosine similarities between the attentions of every pair of options, $L_2$ is the temporal regularization loss for option attentions, and $w_1$ and $w_2$ are the respective weights for these additional objectives. Through their influence on the option attentions, different values for these weights  allow options with the desired characteristics to be obtained. We found that respective values of 4.0 and 2.0 for $w_1$ and $w_2$ resulted in the most diverse options, judged quantitatively (see section \ref{four-rooms-quant-appendix}) and qualitatively (Figure \ref{options_plots}). Further details regarding hyperparameters and reproducibility are given in section \ref{reproducibility}. 

The resulting option usage and their attentions are shown in Figure \ref{options_plots}. The learned options are distinct and specialized in their behavior, and they perform state abstraction by focusing on a subset of the observation vector to perform their specific tasks. The option usage respects the corresponding area of attention, which indicates that the options are typically limited to this area and that their behavior can be reasonably understood from their attentions. AOC also learns stable options and the behavior and usage of options does not vary significantly during the course of training. This stability could be useful for effectively scaling options to  more abstract behaviors. This is in contrast to option-critic (OC), which tends to learn degenerate options that are volatile and continuously change. A qualitative comparison that demonstrates AOC option stability is shown in section \ref{option_stability_section}.

Although AOC additionally needs to learn option attentions, it learns faster than OC, as shown in Figure \ref{training_curves_a}. One possible reason is that in AOC, options specialize to different regions and enable quicker learning, because of smaller overlap between their applicability regions. Furthermore, a comparison between option domination in AOC and OC (see \ref{dominant_option_usage_appendix}) during the course of training indicates that the latter prevents it.

\subsection{Transfer in the four-rooms environment}
We perform two experiments to assess the transfer capability of AOC in the four-rooms domain, both after 30,000 episodes of training. The first is goal transfer, where the goal is moved to a new random location. The second is blocked hallway, where the goal is the same but a random hallway is blocked. To preserve the knowledge learned from the original task, the main network body is frozen, and only the heads and the attentions (in the case of AOC) are updated. AOC transfer I and transfer II respectively represent the scenarios where the weights $w_1$ and $w_2$ are kept unchanged or are set to 0 to give priority to option learning over attention regularization, before learning in the new task. From Figures \ref{training_curves_b} and \ref{training_curves_c}, it can be seen that in spite of the option volatility that aids OC transfer, AOC is faster to recover. In the blocked hallway setting, both variants of AOC perform similarly while OC fails altogether to transfer in multiple runs. This may be explained by the fact that AOC needs to flip only one option policy around, while OC may need to relearn the entire policy for the new task, if the options had already degenerated prior to transfer. Figures \ref{training_curves_b} and \ref{training_curves_c} also show that right after transfer, AOC has a lower decline in performance, compared to OC.

\begin{figure}[!b]
\centering
\includegraphics[scale=0.35]{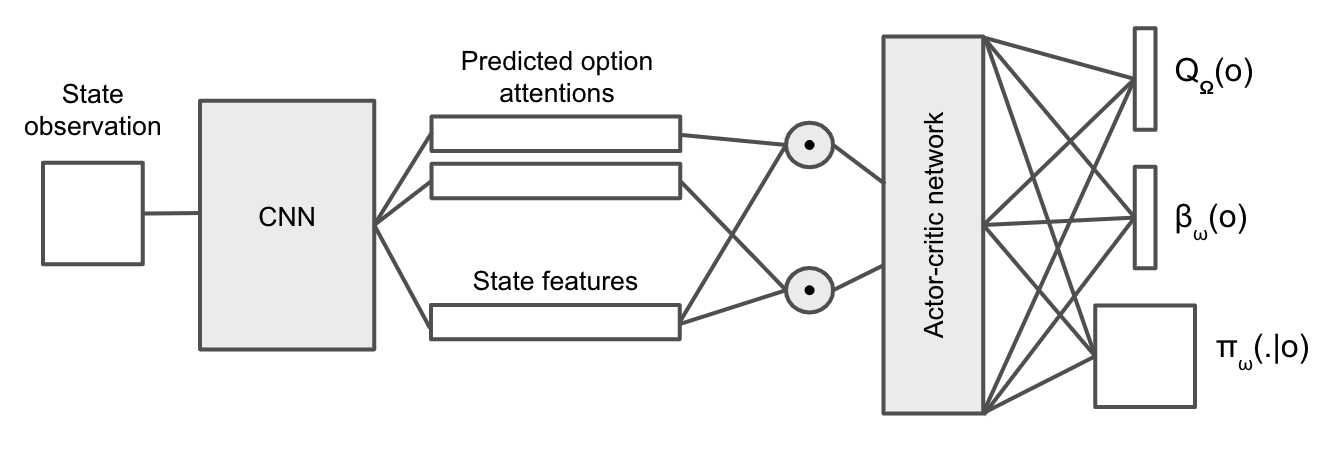}
\caption{The shared network model for AOC in Atari environments. The option attentions are observation dependent. The $\odot$ symbol represents element-wise multiplication.}
\label{atari_high_archi}
\end{figure}

\begin{figure*}[!t]
  \centering
  \subfloat[][Seaquest]{\includegraphics[scale=0.34]{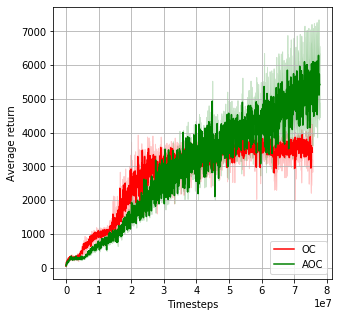} \label{atari_images_a}}
  \subfloat[][Fishing Derby]{\includegraphics[scale=0.34]{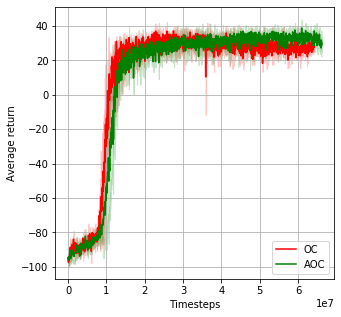} \label{atari_images_b}}
  \subfloat[][Krull]{\includegraphics[scale=0.34]{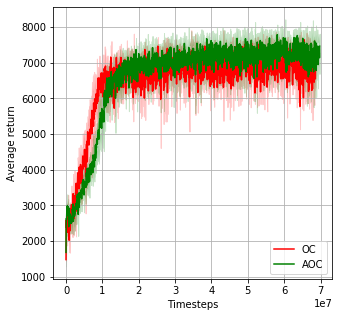} \label{atari_images_c}}
  \subfloat[][Pooyan]{\includegraphics[scale=0.34]{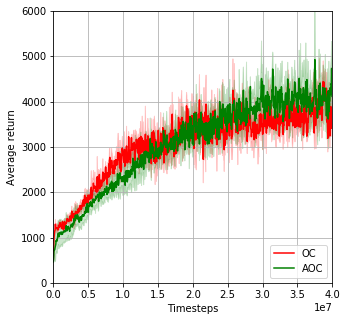} \label{atari_images_d}}
\caption{Learning curves (averaged over 5 seeds) in the Arcade Learning Environment.}
\label{atari_aoc_high_training_curves}
\end{figure*}

\begin{figure*}[!t]
\centering
\includegraphics[scale=0.35]{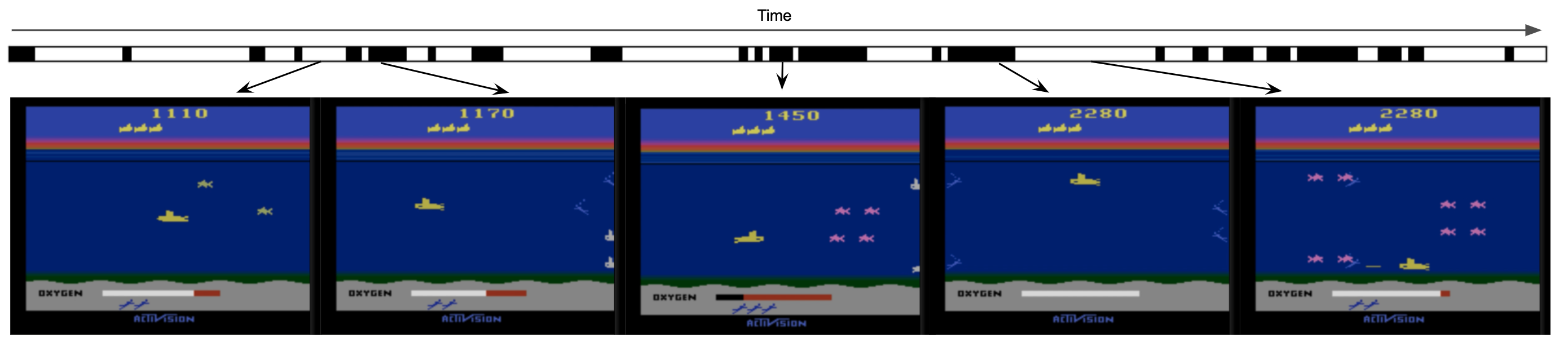}
\caption{An example of AOC option usage in Seaquest over time. Option 0 (shown in white) is used about 70\% of the time and appears to focus on the fish and diver sprites near the middle. Option 1 (shown in black) is used about 30\% of the time and seems to focus on the oxygen bar and new sprites coming onto the screen along the edges.}
\label{seaquest_option_usage}
\end{figure*}

From another perspective, upon transfer, option-critic completely relearns the options. Figure \ref{options_transfer} shows a specific instance of transfer. Comparing  Figure \ref{options_plots_a} with Figures \ref{options_transfer_a} and \ref{options_transfer_b} shows that there is little similarity between the option behavior before and after transfer with OC. We argue that for options to be beneficial for generalization, they should exhibit similar behavior upon transfer,  so that previously learned behaviors can be leveraged, and so that options can be efficiently composed through abstract decisions. AOC exhibits this quality. A comparison of Figures \ref{options_plots_b} and \ref{options_plots_c} with Figures \ref{options_transfer_c} to \ref{options_transfer_f} shows that option attentions remain fixed, indicating that each option remains in its assigned space, and that the option behavior remains relatively consistent upon transfer. These properties of selective transfer and reusability with AOC lead to modular options that speedup transfer.

\subsection{Arcade Learning Environment}
\label{atari}

We now illustrate the performance of AOC in the Arcade Learning Environment \cite{bellemare13}. We use 2 options with a discount factor of 0.99. The input observation $s$ is a stack of 4 frames. The option attentions $h_{\omega,\phi}$ are state dependent and are learned with a convolutional neural network (CNN). These option attentions are then applied to the high level features that are also obtained from the CNN. Each option's attention has the same dimensions as the flattened feature vector. The option policies, values and terminations are learned with a shared-parameter deep network, as in option-critic. The architecture is shown in Figure \ref{atari_high_archi}.

Apart from maximizing the total expected return, the attentions are constrained to exhibit some desired characteristics. Attention diversity is enforced by maximizing the L1 norm between the option attentions and attention sparsity is incentivized by applying an L1 regularizer on the option attentions to penalize non-zero values. Frame stacking implicitly enforces temporal regularization between attentions of successive frames, so we do not specially account for this. Thus, for the Atari domain, the $L$ term in Algorithm \ref{alg:aoc} is enforced by adding $w_{1}L_{1}+w_{2}L_{2}$ to the network loss function, where $L_1$ and $L_2$ are the losses for attention diversity and sparsity respectively. $w_{1}$ and $w_{2}$, represent their respective weights. More details regarding hyperparameters are provided in section \ref{reproducibility}.

For training in the ALE environment, we found that the values 500 and 0.02 for the weights $w_{1}$ and $w_{2}$ respectively, resulted in diverse attentions and good performance. Figure \ref{atari_aoc_high_training_curves} shows the training curves, which indicate that AOC achieves a similar sample complexity (and slightly better final performance) compared to OC, despite also having to learn the state-dependent attention mechanism. We reason that learning the attentions enables options to specialize early on in the training process, and hence speeds up training, despite having more parameters to learn. Additionally, attention sparsity allows the options to focus on only the most important high-level features, which further speeds up learning. For these values of $w_{1}$ and $w_{2}$,  each option focuses on approximately 20\% of the high level features. Figure \ref{seaquest_option_usage} shows the option usage for Seaquest. The options have learned to distinctly specialize their behavior (by learning to focus on different abstract high-level features), and remain relatively balanced in their usage, unlike OC, whose options eventually degenerate as training progresses, since diversity is not incentivized. Thus, AOC allows learning diverse options in complex environments as well. Additional Atari results are shown in section \ref{atari-results-appendix}.

\section{Related work}
\label{related}

There has been extensive research on the benefit of temporal abstraction for reinforcement learning \cite{parr98, dayan93, dietterich00, mcgovern01, stolle03, mann14, kulkarni16}. Specific to the options framework \cite{Sutton99, Bacon17}, there have been several recent approaches to incentivize learned options to be diverse \cite{eysenbach18}, temporally extended \cite{Harb18}, more abstract \cite{riemer18}, and useful to plan \cite{Harutyunyan19} and explore \cite{jinnai19} with. 

The interest option-critic method \cite{Khetarpal20} provides a gradient-based approach towards learning where to initiate options by replacing initiation sets with differentiable interest functions. However, the initialization of the interest functions is biased towards all options being available everywhere. In contrast, our AOC approach is completely end-to-end and does not require any special initializations, and in effect, is able to learn distinct areas where options can be initialized and remain active. 

Deep skill chaining \cite{bagaria20} is another approach that relaxes the assumption of universal option use. This method learns a chain of options by backtracking from the goal and ensuring that the learned initiation set of one option overlaps with the termination of the preceding option. Although each option performs state abstraction, the resulting options are highly dependent on the given task and must be relearned upon transfer. Furthermore, results were mostly confined to navigation-based tasks.  

The MAXQ \cite{dietterich00} approach towards hierarchical reinforcement learning decomposes the value function of the target MDP into value functions of smaller MDPs. Although this decomposition creates an opportunity to perform state abstraction, the overall approach is based on the heavy assumption that the  subtasks  are specified beforehand.  

\section{Conclusion}
To the best of our knowledge, our method is the first to combine temporal and state abstraction in a flexible, end-to-end, gradient based approach. It results in learned options that are diverse, stable, interpretable, reusable and transferable. We demonstrate that the addition of an attention mechanism prevents option degeneracy, a major, long-standing problem in option discovery, and also relaxes the assumption of universal option availability. It also provides an intuitive method to control the characteristics of the learned options.

From the continual learning perspective, an interesting future direction is to meta-learn the attentions and options across a range of tasks from the same environment. This could lead to faster transfer, while keeping the existing benefits of our approach. From the perspective of model-based reinforcement learning, predictive approaches with option attentions could allow for efficient long-horizon planning, by predicting option activation through predicted attentions. Lastly, the approach we have presented is versatile and can be applied to many existing option discovery methods. We leave such avenues of possible combination as future work. 

\bibliography{main}

\onecolumn
\appendix
\section*{Appendix}

\section{Other four-rooms experiments}
\subsection{Comparison of option stability between AOC and OC}
\label{option_stability_section}
\begin{figure*}[!h]
    \centering
    \begin{subfigure}{0.05\linewidth}
      \caption{}
      \label{options_stability_a}
    \end{subfigure}
    \begin{subfigure}{0.54\linewidth}
        \includegraphics[width=\linewidth]{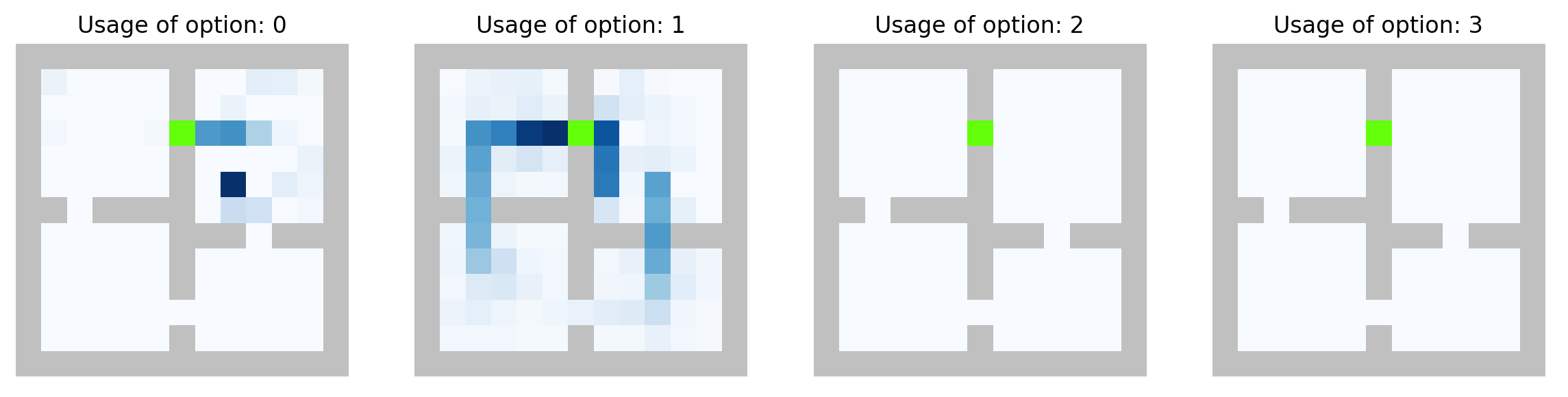}
    \end{subfigure}

    \begin{subfigure}{0.05\linewidth}
      \caption{}
      \label{options_stability_b}
    \end{subfigure}
    \begin{subfigure}{0.54\linewidth}
        \includegraphics[width=\linewidth]{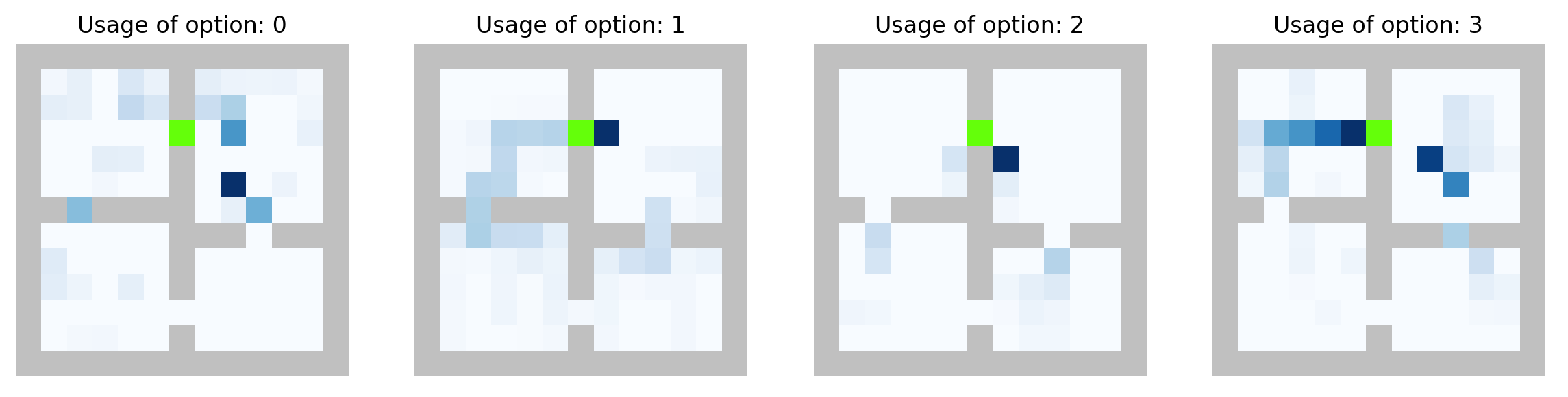}
    \end{subfigure}

    \begin{subfigure}{0.05\linewidth}
      \caption{}
      \label{options_stability_c}
    \end{subfigure}
    \begin{subfigure}{0.54\textwidth}
        \includegraphics[width=\textwidth]{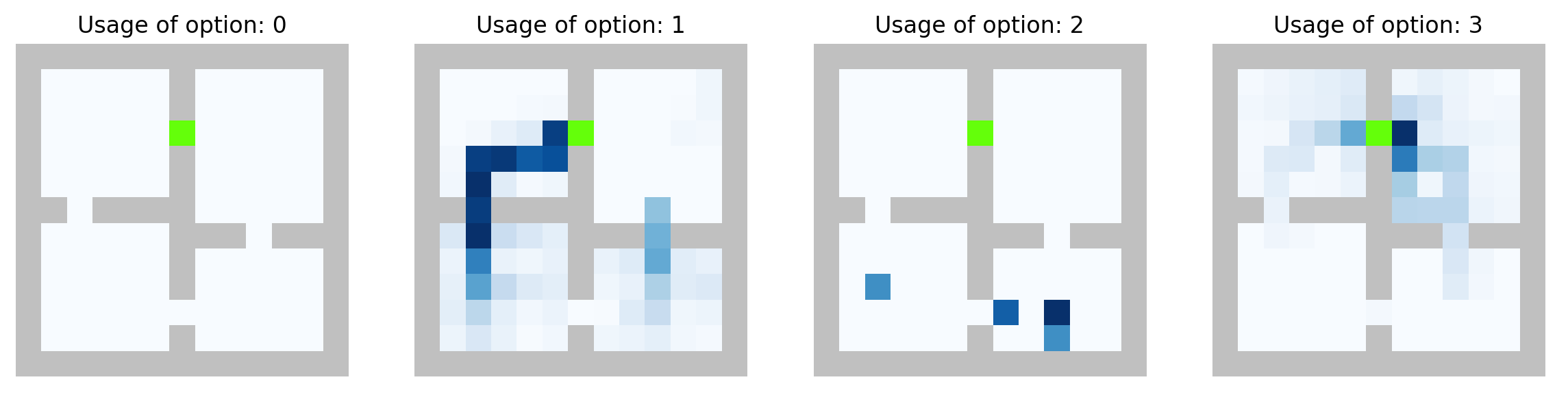}
    \end{subfigure}
    
    \begin{subfigure}{0.05\linewidth}
      \caption{}
      \label{options_stability_d}
    \end{subfigure}
    \begin{subfigure}{0.54\linewidth}
        \includegraphics[width=\linewidth]{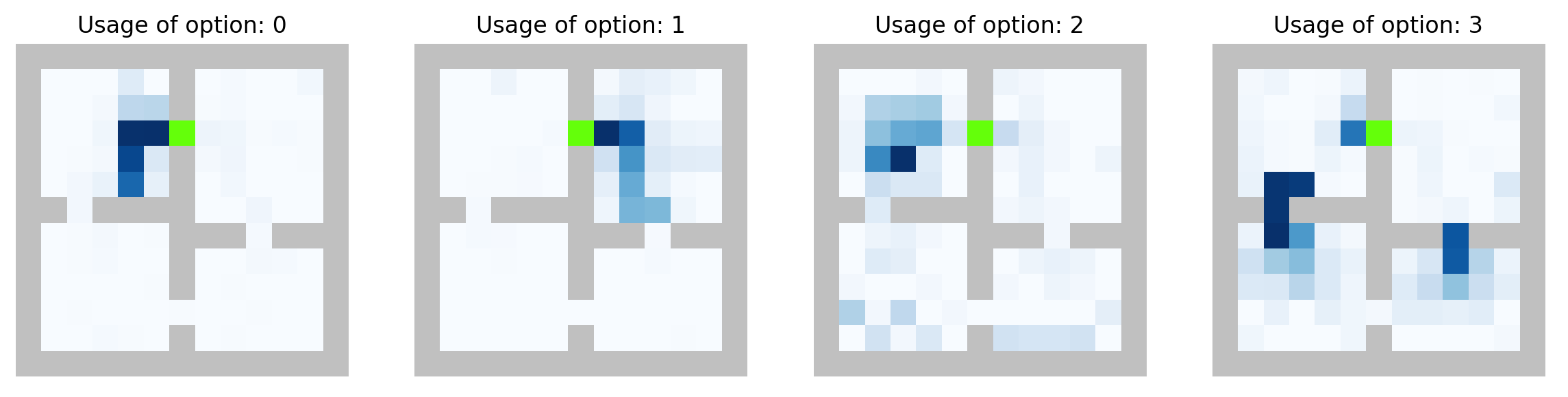}
    \end{subfigure}

    \begin{subfigure}{0.05\linewidth}
      \caption{}
      \label{options_stability_e}
    \end{subfigure}
    \begin{subfigure}{0.54\linewidth}
        \includegraphics[width=\linewidth]{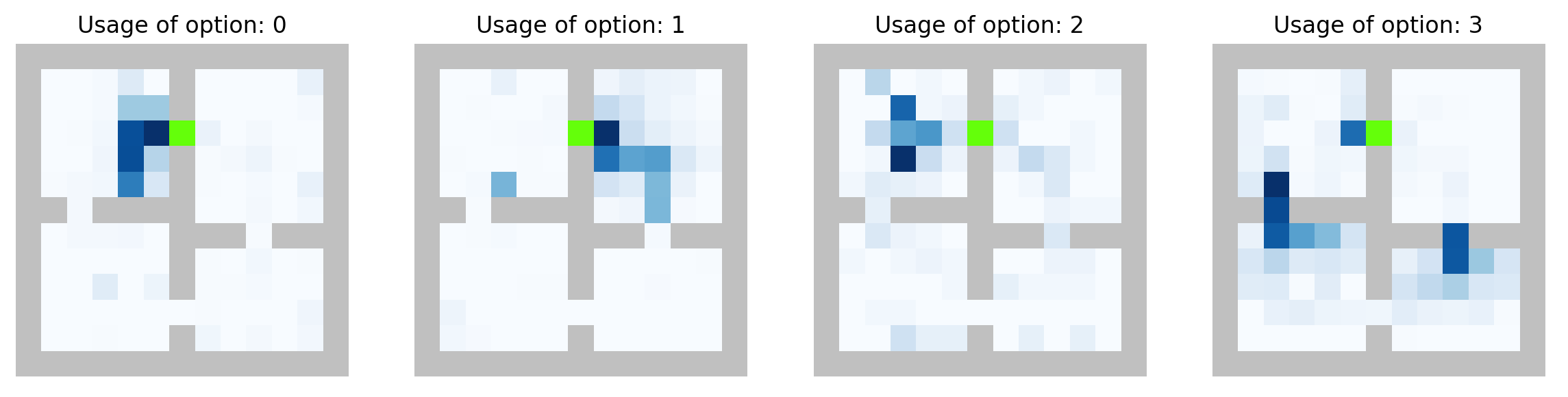}
    \end{subfigure}

    \begin{subfigure}{0.05\linewidth}
      \caption{}
      \label{options_stability_f}
    \end{subfigure}
    \begin{subfigure}{0.54\textwidth}
        \includegraphics[width=\textwidth]{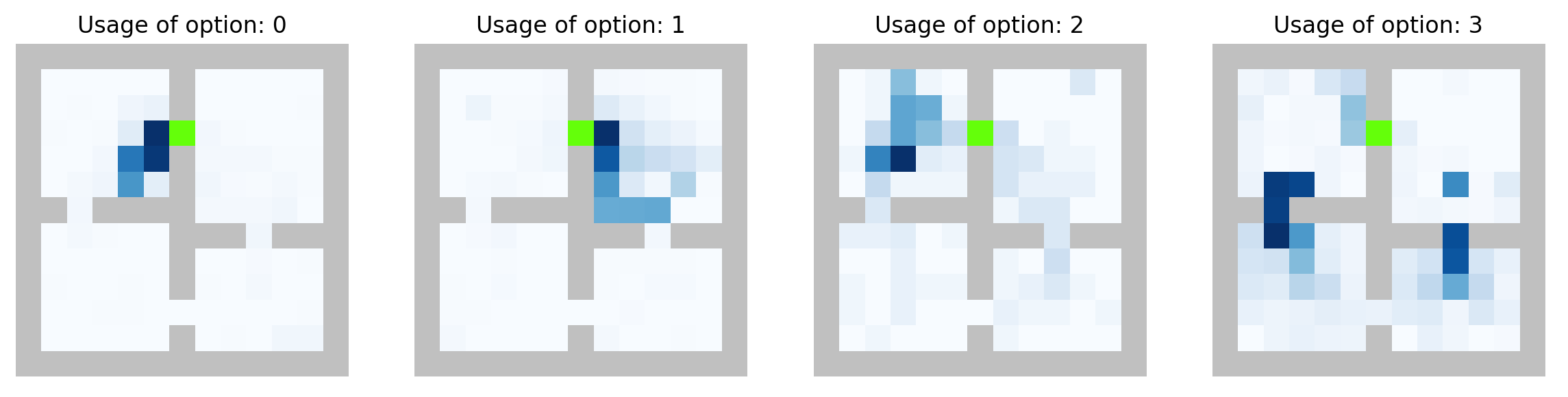}
    \end{subfigure}
    
    \caption{\textbf{(a) to (c):} even after convergence, options learned with OC are volatile and continue to change frequently. \textbf{(d) to (f):} AOC learns more stable options which continue to exhibit similar behavior. In the snapshots of the options above, for both OC and AOC, 100,000 frames of training has been performed between successive rows. The goal is the north hallway, shown in green.}
    \label{options_stability}
\end{figure*}

For a given task, even after convergence, the options learned with OC continue to change significantly in behavior. However, with AOC, the attentions remain fixed after convergence and ensure that the options more stable after convergence. A comparison between OC option volatility and AOC option stability is shown in Figure \ref{options_stability}. Option stability is important for a wide variety of reasons. For example, stable low level options can be composed into higher levels of more complex behavior. Scaling to more abstract levels of behavior may not be efficient if the low level options are volatile. Another benefit of option stability is better model-based planning with options. If options are volatile, it makes it much harder for the agent to learn accurate models that predict where options will terminate if executed. Stable options make it much more likely that a certain option will terminate in a specific region of state space, and therefore enable more accurate planning.

\subsection{Comparison of dominant option usage in AOC and OC}
\label{dominant_option_usage_appendix}

\begin{figure}[!h]
\begin{center}
\centerline{\includegraphics[scale=0.42]{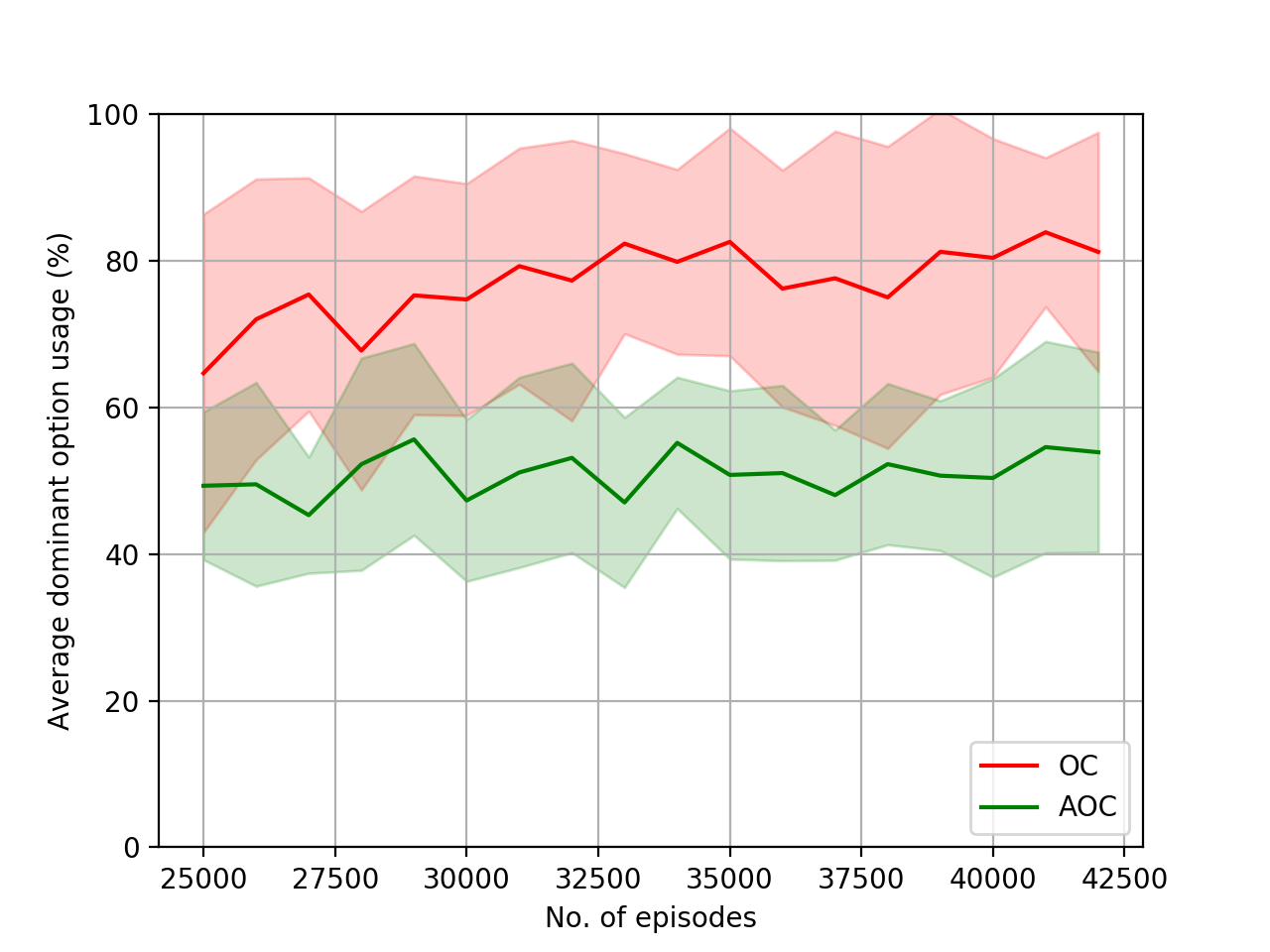}}
\caption{Comparison of average usage of the dominant option in the four-rooms domain.}
\label{max_option_usage}
\end{center}
\end{figure}

A comparison of the usage of the dominant option in AOC and OC is shown in Figure \ref{max_option_usage}.
At each training checkpoint, the dominant option usage is averaged over 50 test episodes for each of the 10 independent training runs. The shaded region represents 1 standard deviation.

\subsection{Hardcoded option attentions}
\label{hardcoded_attention}
\begin{figure}[!h]
\begin{center}
\centerline{\includegraphics[scale=0.45]{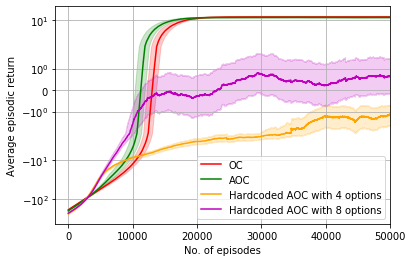}}
\caption{Comparison of OC, AOC, and AOC with hardcoded attentions}
\label{hardcoded_attentions}
\end{center}
\end{figure}

In the case of hardcoded attention where each option's attention is manually limited to one specific and distinct room (i.e. 1 for all states inside the room and 0 elsewhere), slower learning is observed. This is likely because hardcoding attentions de facto removes option choice from the agent, and requires all options to be optimal to get good performance. When we tried hardcoded attention with 8 options (2 per room), we got better performance, but still significantly slower than AOC and OC. Figure \ref{hardcoded_attentions} shows the comparison of the learning curves. Each curve is averaged over 10 runs and the shaded region indicates 0.25 standard deviation.

\subsection{Quantitative measures for four-rooms options and attentions}
\label{four-rooms-quant-appendix}
All of the following quantitative measures are averaged over 10 independent runs with different goal locations.

\subsubsection{Quantitative measure for attention diversity}
\label{attention_diversity_measure_appendix}
After training, the argmax operation applied on the option dimension across the attention maps gives the option with most attention for each state in the environment. Let the option which has the highest attention in most states be termed the most attentive option and let the ratio of its number of highest attention states to total states be called most attentive option coverage. Similarly, let the option which has the highest attention in least states be termed the least attentive option and let the ratio of its number of highest attention states to total states be called least attentive option coverage. The closer both the least and most attentive option coverages are to 25\% (in the case of 4 options), the more diverse the attentions. When the weights $w_1$ and $w_2$ are 4.0 and 2.0 respectively  (which we found to be the most optimal), least attentive option coverage = 9.49\% and most attentive option coverage = 48.14\%. These values indicate that each option has a non-zero area where it is most attentive. 

\subsubsection{Quantitative measure for attention overlap}
\label{attention_overlap_measure_appendix}

After training, let the matrix of maximum attention values for each state (across options) be termed as $max\_attention\_matrix$. Let the matrix of next maximum (2nd highest) attention values for each state (across options) be termed as $second\_max\_attention\_matrix$. Let the difference between these two matrices be called $diff$. Then, a measure of the percentage of state space area where only one option dominates in attention to can be calculated as $sum((diff>0.3)\&\&(second\_max\_attention\_matrix<0.05))*100/total\_states$. Here, $\&\&$ denotes the element-wise logical and operation. This measure calculates the percentage of area where there is minimal competition among option attentions and there is clearly only one option's attention for each state in this area. The higher this measure is, the better. When the weights $w_1$ and $w_2$ are 4.0 and 2.0 respectively, this measure was 68.39\%. For the remaining 31.61\% of the area, it was usually observed to be the case that 2 options' attentions competed for this area (note that this also includes cases where the difference in option attentions is very high i.e. 0.5 or greater but where the second highest option attention was non negligible like 0.15).

\subsubsection{Quantitative measures of variance in option usage}
\label{option_usage_variance_appendix}
The mean option usage for both AOC and OC is near 0.25 for each option (option domination balances out across runs in OC). The standard deviation of option usages for AOC and OC are respectively [0.17, 0.18, 0.18, 0.19] and [0.27, 0.33, 0.31, 0.30] i.e. OC has 3 to 4 times more variance.

\subsubsection{Quantitative measure of consistency between option attentions and usage}
\label{option_usage_attention_consistency_appendix}
The probability that an option is executed when its corresponding attention in a state is $<$0.01 is only 0.17. This indicates that option usage is largely consistent with the corresponding option attentions.

\begin{figure*}[!h]
    \centering
    \begin{subfigure}{0.05\linewidth}
      \caption{}
      \label{overlapping_options_a}
    \end{subfigure}
    \begin{subfigure}{0.6\linewidth}
        \includegraphics[scale=0.32]{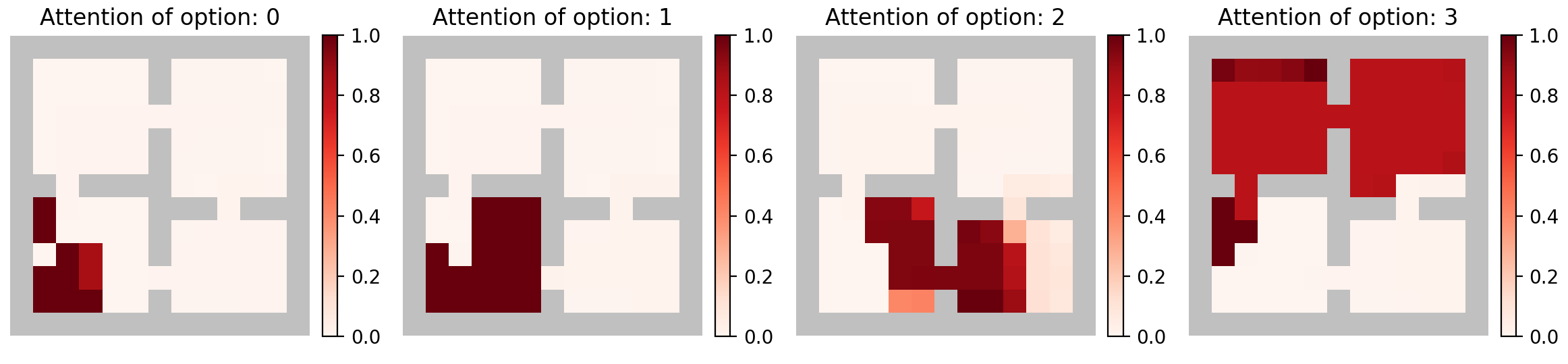}
    \end{subfigure}

    \begin{subfigure}{0.05\linewidth}
      \caption{}
      \label{overlapping_options_b}
    \end{subfigure}
    \begin{subfigure}{0.6\linewidth}
        \includegraphics[scale=0.32]{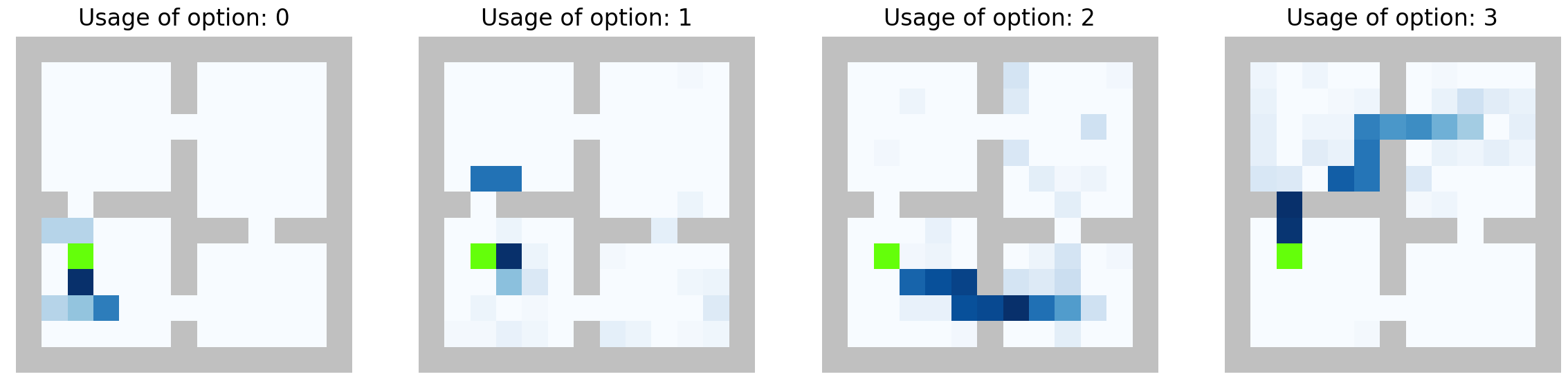}
    \end{subfigure}
    
    \caption{When multiple options have significant overlapping attentions in a state, any of these options may be executed. The goal is shown in green.}
    \label{overlapping_options}
\end{figure*}

It should be noted that in the cases where multiple options have significant non-zero attentions in a state, it can be expected that any of these options may be executed. For example, Figure \ref{overlapping_options} shows the case where multiple options attend to states in the bottom right room. In this case, there is some overlap between the usage of the options that have high attention in these states. Usage in other rooms is still quite distinct.

\subsection{Ablation study on parameters in 4 rooms}
\label{ablation-study-4rooms-params}
\begin{figure*}[!h]
    \centering
    \begin{subfigure}{0.05\linewidth}
      \caption{}
      \label{ablation-study-4rooms-params_images_a}
    \end{subfigure}
    \begin{subfigure}{0.6\linewidth}
        \includegraphics[scale=0.32]{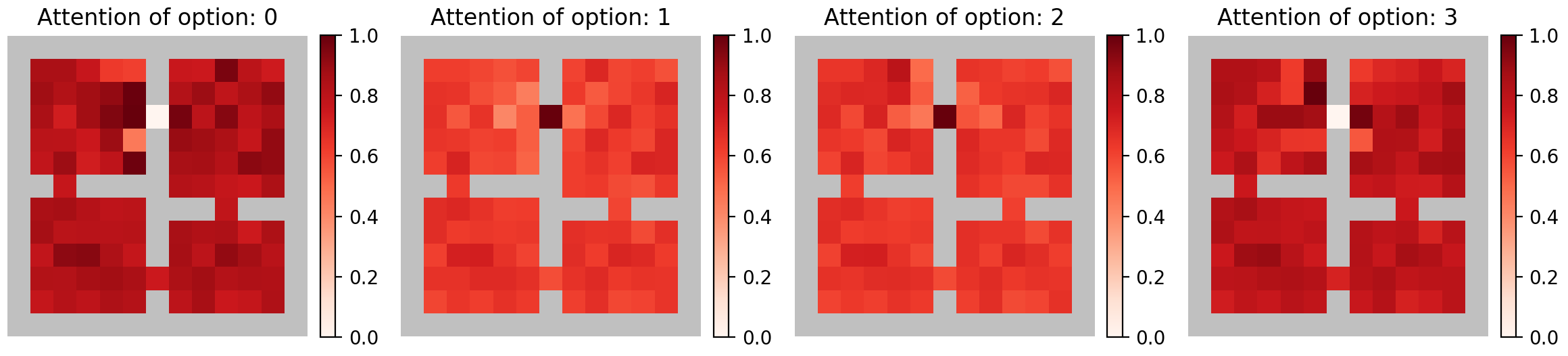} 
    \end{subfigure}

    \begin{subfigure}{0.05\linewidth}
      \caption{}
      \label{ablation-study-4rooms-params_images_b}
    \end{subfigure}
    \begin{subfigure}{0.6\linewidth}
        \includegraphics[scale=0.32]{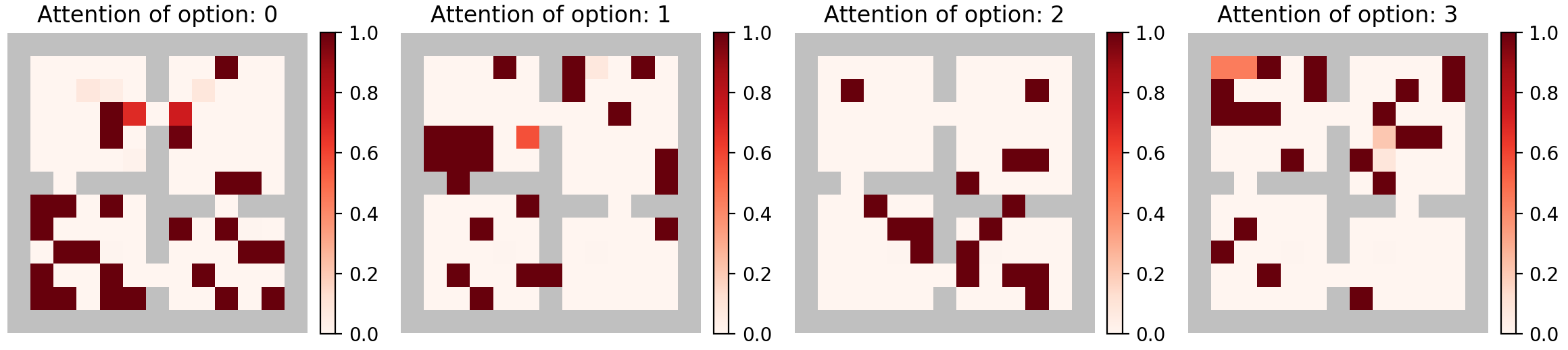}
    \end{subfigure}
    
    \caption{\textbf{(a)} Example of option attentions that are learned when the diversity weight $w_1$ is too low \textbf{(b)} Example of option attentions that are learned when diversity weight $w_1$ is too high and temporal regularization weight $w_2$ is too low.}
    \label{ablation-study-4rooms-params_images}
\end{figure*}

Figure \ref{ablation-study-4rooms-params_images} shows the importance of both the attention diversity and temporal regularization objectives towards learning option attentions with the desired characteristics.

\subsection{Ablation study on number of options in 4 rooms}
\label{ablation-study-num_options}
\begin{figure*}[!bh]
    \centering
    \begin{subfigure}{0.04\linewidth}
      \caption{}
      \label{ablation-study-num_options_images_a}
    \end{subfigure}
    \begin{subfigure}{0.35\linewidth}
        \includegraphics[scale=0.4]{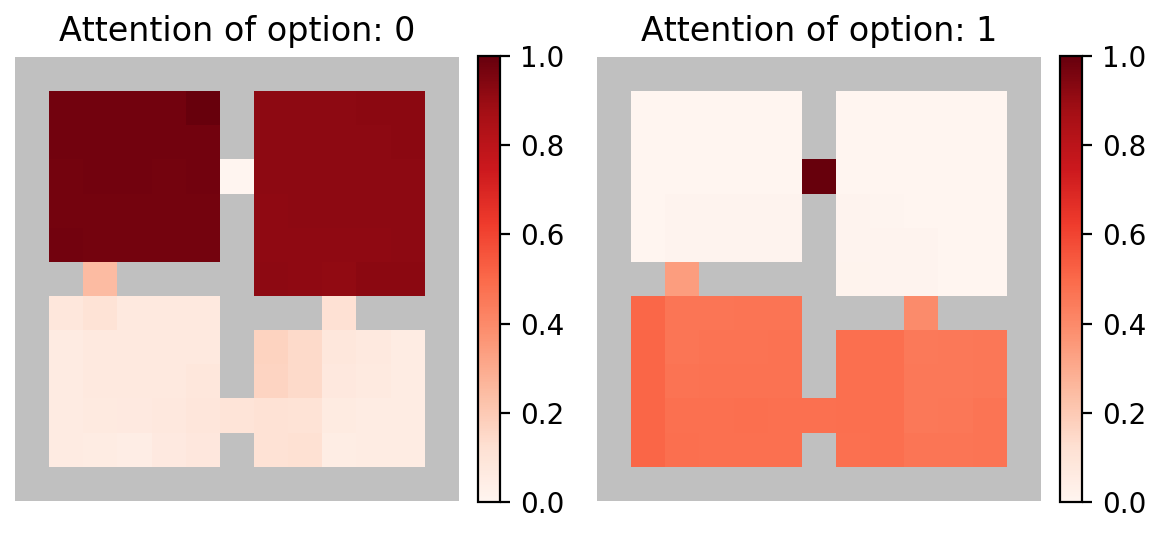} 
    \end{subfigure}
    \begin{subfigure}{0.04\linewidth}
      \caption{}
      \label{ablation-study-num_options_images_b}
    \end{subfigure}
    \begin{subfigure}{0.35\linewidth}
        \includegraphics[scale=0.4]{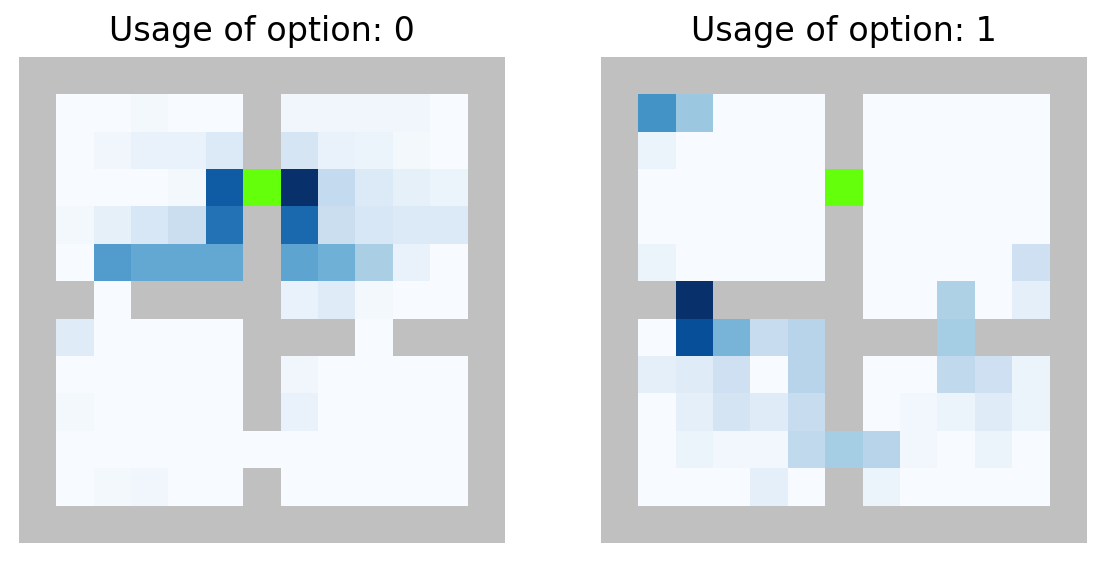}
    \end{subfigure}
    
    \begin{subfigure}{0.19\linewidth}
      \caption{}
      \label{ablation-study-num_options_images_c}
    \end{subfigure}
    \begin{subfigure}{0.55\linewidth}
        \includegraphics[scale=0.46]{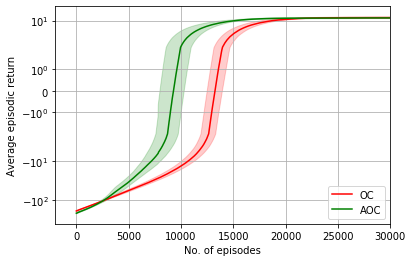}
    \end{subfigure}
    
    \begin{subfigure}{0.09\linewidth}
      \caption{}
      \label{ablation-study-num_options_images_d}
    \end{subfigure}
    \begin{subfigure}{0.85\linewidth}
        \includegraphics[scale=0.49]{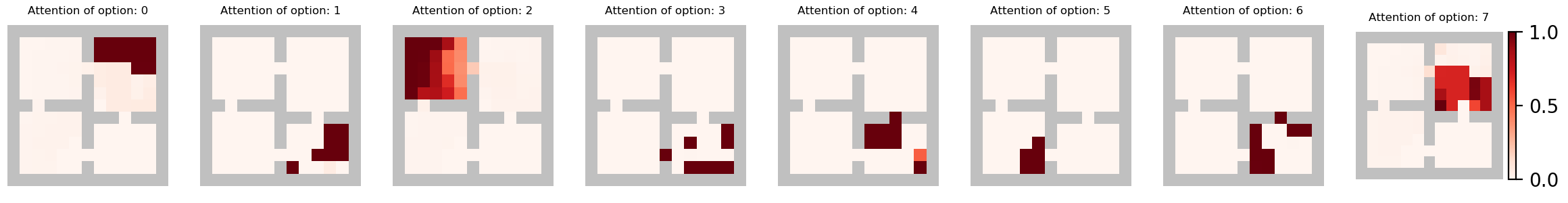} 
    \end{subfigure}
    \begin{subfigure}{0.09\linewidth}
      \caption{}
      \label{ablation-study-num_options_images_e}
    \end{subfigure}
    \begin{subfigure}{0.85\linewidth}
        \includegraphics[scale=0.49]{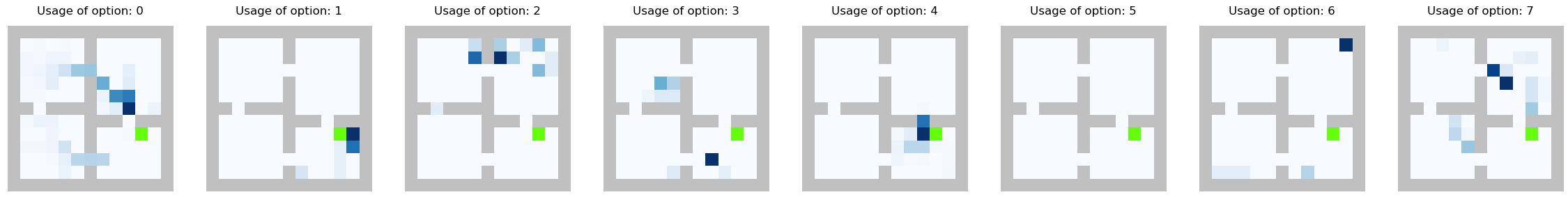}
    \end{subfigure}
    
    \begin{subfigure}{0.19\linewidth}
      \caption{}
      \label{ablation-study-num_options_images_f}
    \end{subfigure}
    \begin{subfigure}{0.55\linewidth}
        \includegraphics[scale=0.46]{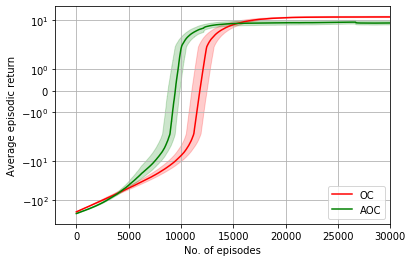}
    \end{subfigure}
    
    \caption{Example of \textbf{(a)} attentions \textbf{(b)} options and \textbf{(c)} training curves when the number of options is 2. Option usage is relatively balanced at 67.2\%, and 32.8\% respectively. The \textbf{(d)} attentions \textbf{(e)} options and \textbf{(f)} training curves when the number of options is 8. Option usage is 44\%, 3\%, 4\%, 5\%, 26\%, 0\%, 2\% and 16\% respectively.}
\label{ablation-study-num_options_images}
\end{figure*}

Figure \ref{ablation-study-num_options_images} shows the effect of the number of options on AOC and the attentions that are learned, and demonstrates that AOC works satisfactorily well for different numbers of options. The same hyperparameter values of 4.0 and 2.0 for $w_1$ and $w_2$ were used. The benefit of AOC with respect to learning speed is even more apparent with 2 options (Figure \ref{ablation-study-num_options_images_c}) than with 4 options (Figure \ref{training_curves_a}).

It should be noted that in the case with 8 options, since AOC motivates option diversity, it can sometimes spread the options too thin and make them occupy only a small area. This tends to slow down learning since many options must be learned well. Some runs (out of the total 10) did not converge in the stipulated training time when there were 8 options and these runs were excluded from Figure \ref{ablation-study-num_options_images_f}, so this figure does not represent a fully fair comparison between AOC and OC when the number of options are high. This problem might be possibly avoided by tuning the weights $w_1$ and $w_2$ specific to the 8 options case, which we did not do. OC does not suffer from this problem when the number of options are high because in such a case, a select few options learn to dominate and the remaining options are ignored.

\section{Other atari experiments}
\label{atari-results-appendix}

\subsection{Atari pixel-level option attentions}
\label{atari-pixel-level-option-attentions}

\begin{figure}[!h]
\begin{center}
\centerline{\includegraphics[scale=0.44]{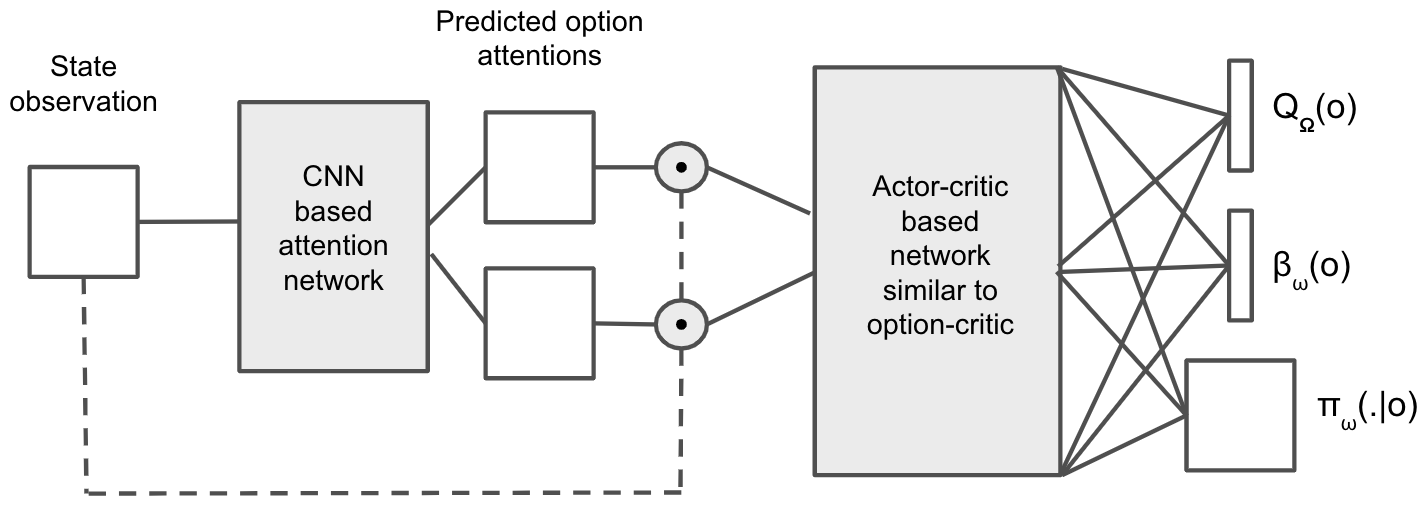}}
\caption{Architecture of AOC for atari with attentions applied at the pixel level. The attentions are observation dependent. $\odot$ denotes element-wise multiplication.}
\label{atari-archi-pixel}
\end{center}
\end{figure}

We also experiment with another approach of AOC in which the attentions are learned and applied at the pixel level, instead of the feature level. The option attentions $h_{\omega,\phi}$ are state dependent and are learned with another convolutional neural network. Each option's attention has the same dimensions as a single frame, and is shared across all frames in the input stack. These learned attentions are then applied to the observation frames, and the result is used to predict the option value, intra-option policy and termination. The architecture is shown in Figure \ref{atari-archi-pixel}.

\begin{figure*}[!b]
  \centering
  \subfloat[][Learning curves]{\includegraphics[scale=0.31]{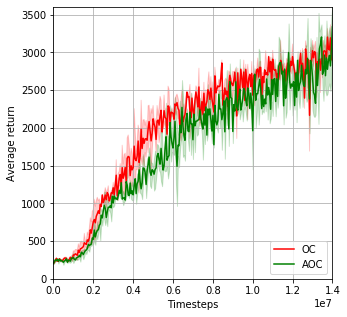} \label{atari_pixel_images_a}}
  \subfloat[][Game frame]{\includegraphics[scale=0.34]{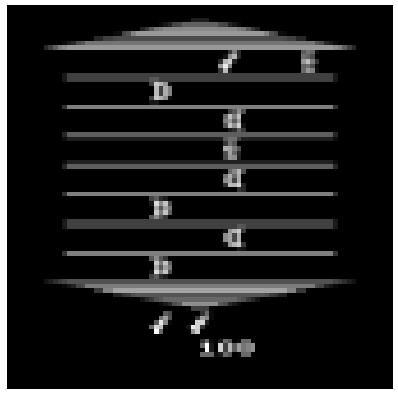} \label{atari_pixel_images_b}}
  \subfloat[][Attention of Option 0]{\includegraphics[scale=0.34]{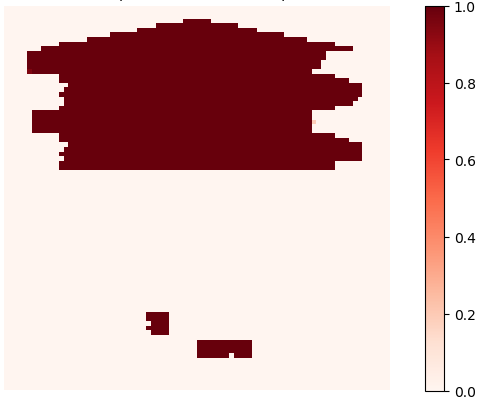} \label{atari_pixel_images_c}}
  \subfloat[][Attention of Option 1]{\includegraphics[scale=0.34]{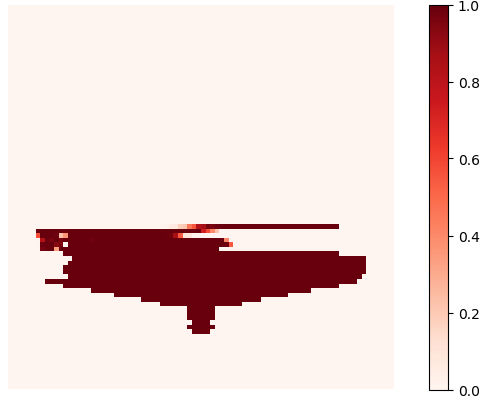} \label{atari_pixel_images_d}}
\caption{AOC atari results for the pixel level attention}
\label{atari_pixel_images}
\end{figure*}

Apart from maximizing the total expected return, the attentions are constrained to exhibit some desired characteristics. Attention diversity is enforced by maximizing the L1 norm between the object attentions of the options and attention sparsity is incentivized by penalizing non-zero attentions for the background. Lastly, attention regularity is promoted between object pixels by penalizing frequent changes in their attention values. The objects and background are identified by finding the connected components in the observation (Figure \ref{atari_pixel_images_b}). Thus, for this variant of atari AOC where the attentions are applied at the pixel level, the $L$ term in Algorithm \ref{alg:aoc} is enforced by adding $w_{1}L_{1}+w_{2}L_{2}+w_{3}L_{3}+w_{4}L_{4}$ to the network loss function, where $L_1$, $L_2$, $L_3$ are the losses for attention diversity, sparsity and regularity respectively. The additional regularizer $L_4$ is added to prevent an option's attention from collapsing to zeros. $w_{1}$, $w_{2}$, $w_{3}$ and $w_{4}$ represent their respective weights.

For training in the Asterix environment, we found that the values 5000, 0.01, 100, and 1 for the weights $w_{1}$, $w_{2}$, $w_{3}$ and $w_{4}$ respectively, resulted in diverse attentions and good performance. Figure \ref{atari_pixel_images} shows the performance and learned option attentions. Figure \ref{atari_pixel_images_a} shows that a similar sample complexity is achieved, compared to OC. Figures \ref{atari_pixel_images_c} and \ref{atari_pixel_images_d} show the resulting option attentions and indicate that option 0 and option 1 have respectively specialized to behaviors pertaining to the main sprite's position in the upper or lower half of the frame.

The main advantage of applying attention at the pixel level (instead of the feature level) is the interpretability. The option attentions clearly indicate where the options focus. However, pixel level attentions also have disadvantages. They require more hyperparameters that additionally need significant tuning to obtain the desired option characteristics. Additionally, the performance seems similar to OC, but not better. Lastly, the option attentions may end up being spatially confined and might be forced to share pixel space on each and every frame, even when this is not needed. Applying attentions to higher level features does not cause this problem and a single option can end up specializing fully to a specific frame layout.    

\section{Reproducibility and training details}
\label{reproducibility}

The models are implemented in PyTorch and experiments were run on an NVIDIA V100 SXM2 with 16GB RAM.

\subsection{Four-rooms environment}
For all experiments in the four-rooms domain, we use the following option learning model for both AOC and the OC baseline: a 3-layer neural (layerwise with 60 and 200 neurons each with ReLU activation, followed by 3 separate heads for option values, intra-option policies and option terminations) with fully-connected branches for option values, intra-option policies (with softmax function) and the option terminations (with sigmoid function). The parameters used for both AOC and baseline OC (after a hyperparameter search) are shown in Table \ref{four-rooms-table}.

\begin{table}[h!]
\begin{center}
\begin{small}
\begin{sc}
\begin{tabular}{lcccr}
\toprule
Parameter & Value \\
\midrule
Number of workers    & 5 \\
Gamma ($\gamma$)    & 0.99 \\
Number of options    & 4 \\
Optimizer    & RMSprop \\
Learning rate    & $10^{-3}$ \\
Option exploration    & Linear($10^{0}$, $10^{-1}$, $10^5$) \\
Entropy    & Linear($10^{2}$, $10^{-1}$, $10^5$) \\
Rollout length    & 5 \\
\bottomrule
\end{tabular}
\caption{Hyperparameters for four-rooms}
\label{four-rooms-table}
\end{sc}
\end{small}
\end{center}
\end{table}

\begin{table}[h!]
\begin{center}
\begin{small}
\begin{sc}
\begin{tabular}{lcccr}
\toprule
Layer & in-channels & out-channels & kernel-size & stride  \\
\midrule
conv1    & - & 32 & 8 & 4 \\
conv2    & 32 & 64 & 4 & 2 \\
conv3    & 64 & 64 & 3 & 1 \\
fc1    & $7\times7\times64$ & 512 & - & - \\
\bottomrule
\end{tabular}
\caption{Option learning model for ALE environment}
\label{option-model-atari}
\end{sc}
\end{small}
\end{center}
\end{table}

\begin{table}[!htbp]
\begin{center}
\begin{small}
\begin{sc}
\begin{tabular}{lcccr}
\toprule
Parameter & Value \\
\midrule
Number of workers    & 16 \\
Gamma ($\gamma$)    & 0.99 \\
Number of options    & 2 \\
Optimizer    & RMSprop \\
Learning rate    & $10^{-4}$ \\
Option exploration    & $10^{-1}$ \\
Entropy    & $10^{-2}$ \\
Rollout length    & 5 \\
Framestack    & 4 \\
\bottomrule
\end{tabular}
\caption{Hyperparameters for ALE}
\label{atari-table}
\end{sc}
\end{small}
\end{center}
\end{table}

We performed a grid search across multiple values for $w_1$ and $w_2$, the weights for cosine similarity between the attentions and the temporal regularization loss respectively. The search space for both weights was the range [0, 5.0] in increments of 0.25. The best values (judged according to qualitative attention diversity and quantitative measures explained above) were found to be 4.0 and 2.0 for $w_1$ and $w_2$ respectively. The shaded regions in Figure \ref{training_curves_a} represent 0.5 standard deviation, and 0.25 standard deviation in Figures \ref{training_curves_b} and \ref{training_curves_c}. Note that all learning curves before transfer in the four-rooms domain are averaged over 10 random goal locations. All learning curves after transfer are also averaged over 10 different runs (with new goal locations and blocked hallways as applicable).

\subsection{Arcade Learning Environment}
For experiments in the Arcade Learning Environment, the structure of the CNN part for both AOC and the OC baseline is shown in Table \ref{option-model-atari}. Each convolution layer is followed by ReLU activation. The FC1 layer is followed by fully-connected branches for option values, intra-option policies (with softmax function) and the option terminations (with sigmoid function). For AOC, the same thing happens, but the FC1 layer features are multiplied by the option attentions and then they're passed through the branched heads. The parameters used for both models of AOC and baseline OC (after a hyperparameter search) are shown in Table \ref{atari-table}. The input observation is a grayscale $84\times84\times4$ tensor.

We performed a grid search across multiple values for $w_1$ (weight for attention diversity) and $w_2$ (weight for attention sparsity). The search space for all weights was the range [$10^{-3}$, $10^4$] in near semi-logarithmic increments. The best weight values for all atari environments (tuned individually) were 500.0 and 0.02 for $w_1$ and $w_2$ respectively. Each atari learning curve is an average over 5 random seeds and the shaded region represents 1 standard deviation. 

\end{document}